\documentclass[10pt,twocolumn,letterpaper]{article}

\usepackage[pagenumbers]{cvpr} %

\usepackage[dvipsnames]{xcolor}

\usepackage{epsfig}
\usepackage{graphicx}
\usepackage{comment}
\usepackage{amsmath,amssymb} %
\usepackage{xcolor}
\usepackage[export]{adjustbox}
\usepackage{bbm}

\def\etal{{\em et al. }}
\usepackage[font=small,skip=5pt]{caption}
\usepackage{float}
\usepackage{makecell} %

\usepackage{multirow}
\usepackage{tabularx}
\usepackage{soul}
\usepackage{array}
\usepackage{comment}
\usepackage{outlines}

\usepackage[mathscr]{euscript}
\newcolumntype{L}[1]{>{\raggedright\let\newline\\arraybackslash\hspace{0pt}}m{#1}}
\newcolumntype{C}[1]{>{\centering\let\newline\\arraybackslash\hspace{0pt}}m{#1}}
\newcolumntype{R}[1]{>{\raggedleft\let\newline\\arraybackslash\hspace{0pt}}m{#1}}
\newcolumntype{Y}{>{\centering\arraybackslash}X}
\usepackage{booktabs}
\usepackage{enumitem}

\usepackage{amsmath}

\newcommand{\mat}[1]{\mathbf{#1}}

\newcommand{\vect}[1]{\mathbf{#1}}

\newcommand{\pose}[0]{\boldsymbol{\theta}}
\newcommand{\shape}[0]{\boldsymbol{\beta}}

\definecolor{ForestGreen}{RGB}{14,109,14}

\usepackage{multirow}

\definecolor{cvprblue}{rgb}{0.21,0.49,0.74}
\usepackage[pagebackref,breaklinks,colorlinks,citecolor=cvprblue]{hyperref}

\def\paperID{***} %
\def\confName{CVPR}
\def\confYear{2024}

\title{RHOBIN Challenge: Reconstruction of Human Object Interaction}

\author{Xianghui Xie$^{1,2,3}$ \qquad Xi Wang$^4$ \quad Nikos Athanasiou$^5$ \quad Bharat Lal Bhatnagar$^6$ \quad Chun-Hao P. Huang$^8$ \\ 
\quad Kaichun Mo$^7$ \quad Hao Chen$^9$ \quad Xia Jia$^9$ \quad Zerui Zhang$^{10}$ \quad Liangxian Cui$^{10}$ \quad Xiao Lin$^{10}$ \\ Bingqiao Qian$^{10}$ \quad
Jie Xiao$^{10}$ \quad Wenfei Yang$^{10}$ \quad Hyeongjin Nam$^{11}$ \quad  Daniel Sungho Jung$^{11}$ \\ Kihoon Kim$^{11}$ \quad 
Kyoung Mu Lee$^{11}$ \quad Otmar Hilliges$^{4}$ \quad Gerard Pons-Moll$^{1,2,3}$ \\
\small $^1$University of T\"ubingen \enskip $^2$T\"ubingen AI Center\enskip $^3$Max Planck Institute for Informatics\enskip $^4$ETH Z\"urich\enskip \\
\small \enskip $^5$Max-Planck-Institute for Intelligen Systems \enskip $^6$Meta Reality Labs \enskip $^7$NVIDIA Research\enskip $^8$Adobe \\
\small $^9$NIP3D-ARTeam \enskip $^{10}$University of Science and Technology of China \enskip $^{11}$Seoul National University  \\
}

\begin{document}
\maketitle
\begin{abstract}
Modeling the interaction between humans and objects has been an emerging research direction in recent years. Capturing human-object interaction is however a very challenging task due to heavy occlusion and complex dynamics, which requires understanding not only 3D human pose, and object pose but also the interaction between them. Reconstruction of 3D humans and objects has been two separate research fields in computer vision for a long time. We hence proposed the first RHOBIN challenge: reconstruction of human-object interactions in conjunction with the RHOBIN workshop. It was aimed at bringing the research communities of human and object reconstruction as well as interaction modeling together to discuss techniques and exchange ideas. Our challenge consists of three tracks of 3D reconstruction from monocular RGB images with a focus on dealing with challenging interaction scenarios. 
Our challenge attracted more than 100 participants with more than 300 submissions, indicating the broad interest in the research communities. This paper describes the settings of our challenge and discusses the winning methods of each track in more detail. We observe that the human reconstruction task is becoming mature even under heavy occlusion settings while object pose estimation and joint reconstruction remain challenging tasks. With the growing interest in interaction modeling, we hope this report can provide useful insights and foster future research in this direction. Our workshop website can be found at \href{https://rhobin-challenge.github.io/}{https://rhobin-challenge.github.io/}.

\end{abstract}    
\section{Introduction}
Humans are in constant contact with the world as they move through it and interact with it. The ability to understand human behavior and how they interact with their surroundings has long been desired in academic and industrial settings. This task is however very challenging. It requires not only techniques for modeling human shape, pose, and motion but also methods that reason about the environments we are living in. Reconstructions of 3D humans and objects have been separate fields in computer vision. Only recently, we have seen emerging trends in modeling the interactions between humans and objects~\cite{GrapingField:3DV:2020, hasson19_obman, Tzionas2016capture-handobject, zhang2020phosa, xie22chore, Yi_MOVER_2022, PROX:2019, zhang2022couch, CVPR21HPS}. With the growing interest in interaction modeling, there is an increased demand for bringing together the communities of 3D human reconstruction, object pose estimation as well as interaction modeling to share knowledge and explore the full potential of this shared research interest and its applications.  

Therefore, we propose the first RHOBIN challenge: \textbf{R}econstruction of \textbf{H}uman \textbf{OB}ject \textbf{IN}teraction in conjunction with the CVPR'23 RHOBIN workshop~\cite{rhobin2023}. The challenge was designed to discuss innovative research directions by connecting human modeling, object-oriented learning, and human-object interaction reconstruction, and explore methods that can be used in a broader context. 

Specifically, we use BEHAVE~\cite{bhatnagar22behave} as the benchmark dataset to evaluate different methods. We propose three tracks in the RHOBIN challenge: 1) 3D human reconstruction, 2) 6DoF pose estimation of rigid objects, and 3) Joint reconstruction of humans and objects. All tasks take a single RGB image as input and output 3D human or/and the interacting 3D object. Our challenge has received lots of attention from the research communities, with a total of 100+ participants and 300+ submissions. All best-performing teams have obtained better results than previous state-of-the-art methods~\cite{xie22chore, rong2020frankmocap}. 

After examining the results, we have the following observations: 1) Existing methods can already achieve quite good results on separate human or object reconstruction and it is important to apply data augmentation and model ensemble to boost performance; 2) The joint reconstruction remains challenging and requires further innovation to improve the accuracy. In terms of the winning methods, we observe that estimating the \emph{2D-3D correspondence} is an essential module for all methods. This is represented either as dense correspondence (object pose) or keypoint estimation (human and joint reconstruction). In addition, while direct regression can achieve quite good performance for human or object reconstruction, an additional optimization layer is still required for the joint reconstruction task.

Overall, we can see that significant progress has been made in separate human or object reconstruction. These methods with additional learning tricks can already obtain quite good results on the challenging BEHAVE dataset. However, the progress in the joint reconstruction task is lagging behind, which calls for further innovation to improve both the speed and accuracy of reconstruction methods. 

In the following sections, we will first discuss the challenge settings in \Cref{sec:challenge} and then present the winning methods of each task in more detail in \Cref{sec:methods}.

\section{Challenge Details}\label{sec:challenge}
In this section, we discuss in more detail the settings of our challenge. We start by describing the dataset we used to benchmark different methods and then define the tasks in this challenge. For each task, we briefly review related works and discuss the metrics as well as baselines for evaluation. We then summarize the results of this challenge. 

\subsection{Dataset}
\begin{figure*}[t]
    \centering
    \includegraphics[width=\linewidth]{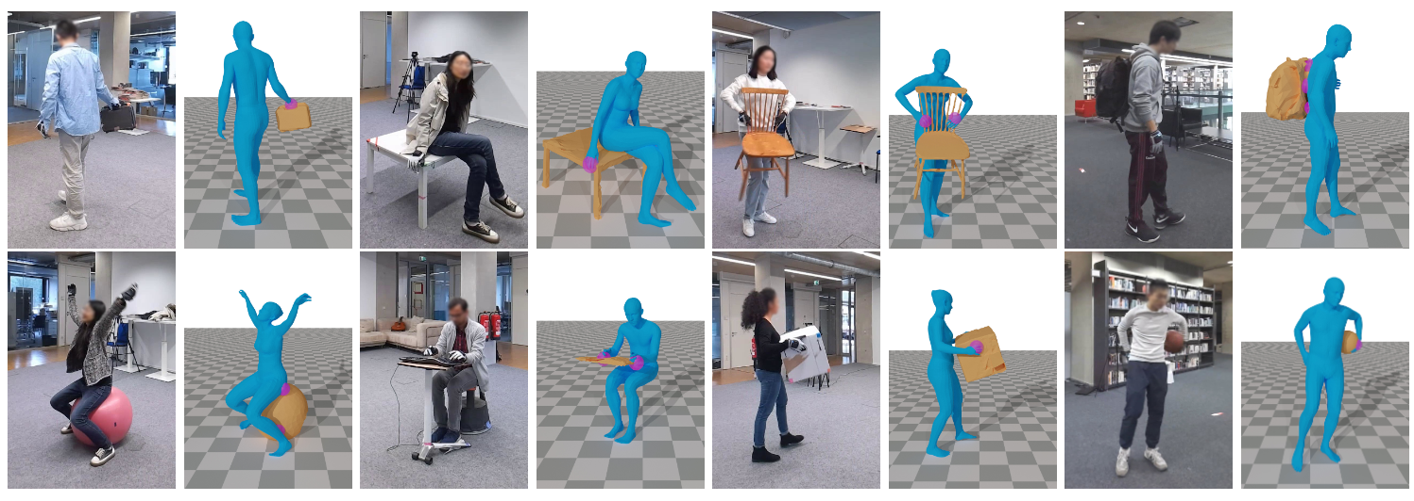}
    \caption{\textbf{Example images and 3D annotations} from the BEHAVE~\cite{bhatnagar22behave} dataset. BEHAVE captures realistic human-object interactions in natural environments.}
    \label{fig:behave-image}
\end{figure*}

We use the publicly available BEHAVE dataset~\cite{bhatnagar22behave} for this challenge. BEHAVE captures humans interacting with objects in natural environments using 4 RGBD cameras. Each RGB image is paired with 3D (pseudo) GT human and object registrations, which makes it possible to benchmark different methods. The scale of the dataset allows the training of large-scale models. Some example images from the dataset are shown in \cref{fig:behave-image}. A comparison of different datasets until the challenge release time is summarized in \cref{tab:datasets}. We can see BEHAVE is the largest dataset so far that captures both human-object interactions in natural environments. 

\begin{table}[ht]
    \centering
    \small
    \begin{tabular}{ c | c | c | c | c}
         Dataset & Modality & 3D Hum. & \# Obj. & \# Frames  \\
         \hline
         3DPW~\cite{vonMarcard2018} &  RGB & Yes & None & 51k \\ 
         Linemod~\cite{linemod} & RGB+D & None & 15 & 18k \\ 
         YCB~\cite{xiang2018posecnn} & RGB & None & 21 & 133k \\ 
         PROX~\cite{PROX:2019} & RGB+D & Yes & Static & 100k \\ 
         InterCap~\cite{huang2022intercap} & RGB+D & Yes & 10 & 67.4k \\ 
         BEHAVE~\cite{bhatnagar22behave} & RGB+D & Yes & 20 & 464.4k \\ 
    \end{tabular}
    \caption{Comparison of different datasets relevant to human and object pose estimation and reconstruction. The BEHAVE dataset is the largest dataset that features both humans and objects and allows for the evaluation of joint human-object reconstructions. }
    \label{tab:datasets}
\end{table}

\subsection{Tasks and Evaluation}
Based on the BEHAVE dataset, we propose three tracks on 3D reconstruction from monocular RGB images, which are detailed next. 

\subsubsection{Object 6DoF pose estimation}
\paragraph{Task definition.} Given a single image $\mat{I}\in \mathbb{R}^{3\times H\times W}$ of an object in interaction and corresponding object mesh template $\mathcal{M}=(\mat{V}, \mat{F})$, the goal is to estimate the rotation $\mat{R}_o\in \text{SO}(3)$ and translation $\vect{t}_o\in \mathbb{R}^3$ parameters of the object in the image w.r.t the template. Following CHORE~\cite{xie22chore}, participants are allowed to use the object mask as additional input. 

\paragraph{Related work.} Object pose estimation has been significantly advanced by deep learning methods in recent years\cite{Wang_2021_GDRN, majcher_shape_nodate, li_deepim_nodate, di_so-pose_2021, peng_pvnet_2019, hu_single-stage_cvpr20, Wang_2019_CVPR_NOCS, liu2022gen6d, sun2022onepose}. These methods can be classified into two-stage or direct regression methods. Two-stage methods first predict keypoints in 2D images~\cite{peng_pvnet_2019} or dense correspondence to a canonical template \cite{Wang_2019_CVPR_NOCS} and then solve a Perspective-n-Point (PnP) problem to find the optimal pose. These methods achieve high accuracy and image alignment but the inference speed is slow. Direct regression methods are thus proposed to use a network to directly predict the object rotation and translation from images \cite{Wang_2021_GDRN, hu_single-stage_cvpr20, di_so-pose_2021}. In particular, GDR-Net~\cite{Wang_2021_GDRN} improves the accuracy by adopting the normalized object coordinate system (NOCS)~\cite{Wang_2019_CVPR_NOCS} and Di \etal~\cite{di_so-pose_2021} leverages self-occlusion to boost performance. These methods allow fast inference speed without losing accuracy. 

\paragraph{Evaluation.} We adopt the error metrics from the BOP challenge~\cite{hodan2018bop}. Given the estimated pose $\hat{\mat{P}}$ and the ground truth pose $\mat{P}$ of an object model $\mat{M}$, we define three pose error functions as below: 
\begin{itemize}
    \item \textbf{MSSD}: $e_\text{MSSD}=\text{min}_{S\in S_\mat{M}}\text{max}_{\vect{x}\in \mat{M}}||\hat{\mat{P}}\vect{x}-\mat{P}\mat{S}\vect{x}||_2$, where $S_\mat{M}$ is a set of manually defined global symmetry transformations for each object. This error is called maximum symmetry-aware surface distance. The maximum distance is relevant for robot manipulation tasks, where this maximum deviation can indicate the chance of a successful grasp. Also, it is less dependent on the sampling of mesh vertices compared to common ADD/AD metrics. 
    \item \textbf{MSPD}: $e_\text{MSSD}=\text{min}_{S\in S_\mat{M}}\text{max}_{\vect{x}\in \mat{M}}||\pi(\hat{\mat{P}}\vect{x}) - \pi(\mat{P}\mat{S}\vect{x})||_2$, where $\pi: \mathbb{R}^3\mapsto \mathbb{R}^2$ denotes a perspective projection. Different from MSSD, this error metric considers the vertices in pixel space and is called maximum symmetry-aware projection distance. 
    \item \textbf{Rotation error (RE)}: $e_\text{RE}=\text{min}_{S\in S_\mat{M}} \text{arccos} ((\text{Tr}(\hat{\mat{R}}\mat{R}S-1)/2))$, where $\hat{\mat{R}}, \mat{R}$ are the estimated and ground truth rotation respectively. It is difficult to obtain absolute translation from monocular RGB input, especially during human-object interaction. However, object orientation remains important because wrong orientation can lead to physically implausible interaction. This rotation error reflects the object orientation estimation accuracy and is relevant for interaction modeling.
\end{itemize}
With these error functions, we can then compute the recall given a threshold for each error. We hence propose the following metrics based on recall thresholds:
\begin{itemize}
    \item \textbf{MSSD-AR}: the average of recall rates calculated for MSSD error threshold ranging from 5\% to 50\% of the object diameter with a step of 5\%.
    \item \textbf{MSPD-AR}: the average of recall rates calculated for MSPD error threshold ranging from 5px to 100px with a step of 5px.
    \item \textbf{RE-AR}: the average of recall rates calculated for RE threshold ranging from 0.05r to 0.5r, where r=40 degrees.
\end{itemize}
We average these recall values to compute a single average recall to rank different methods. 

\paragraph{Baseline.} Previous object pose estimation methods were commonly evaluated using the BOP benchmark~\cite{hodan2018bop}, which does not have human interaction and contains less occlusion. CHORE~\cite{xie22chore} is the first method that predicts object pose on the challenging BEHAVE dataset. We hence use the object reconstruction results from CHORE as the baseline for this track. Its performance is shown in \cref{tab:object-6dof} row 1. Due to heavy occlusions and the handling of human interaction, the average recall is 0.3, leaving a large space for further improvement. 

\subsubsection{Human reconstruction}
\paragraph{Task definition.} Given a single image $\mat{I}\in \mathbb{R}^{3\times H\times W}$ of a person in interaction, the goal is to reconstruct the person in 3D. We use SMPL~\cite{smpl2015loper} as the human representation hence the output is SMPL pose $\pose\in \mathbb{R}^{3+72}$ (including global translation) and shape $\shape\in \mathbb{R}^{10}$ parameters. 

\paragraph{Related work.}
Human reconstruction is a fundamental research field that has been studied for decades. Based on the requirement of a template body shape or not, methods can be categorized into template-based~\cite{rong2020frankmocap, VIBE:CVPR:2020, hmrKanazawa17, li2021hybrik, SMPL-X:2019, kolotouros2019spin, bogo2016smplify} and template-free methods~\cite{pifuSHNMKL19, saito2020pifuhd, li2020monocap}. Template-based methods usually use a statistical body model such as SMPL~\cite{smpl2015loper} to represent the 3D human and then the problem boils down to estimating the parameters of the body model. Along this direction, recent methods follow either optimization based~\cite{bogo2016smplify, SMPL-X:2019} or direct regression~\cite{hmrKanazawa17, rong2020frankmocap, kolotouros2019spin} paradigms. On the other hand, template-free methods do not rely on any predefined body shape, hence can flexibly reconstruct the clothing deformation as well. Template-free methods rely purely on the network to learn the shape prior, which might not be robust under challenging poses. Hence recent works~\cite{xiu2022icon, xiu2023econ} combine SMPL estimation from image with the network priors for the clothing to obtain robust reconstruction under various poses in the wild. 
More recent works~\cite{huang2024tech, liao2023tada} treat the reconstruction as a conditional generation problem and explore pre-trained diffusion models to reconstruct both clothing, hand and head in detail. For a comprehensive review of mesh recovery methods, please refer to~\cite{tian2022hmrsurvey}.

\paragraph{Evaluation.} We use the same evaluation metrics used in 3DPW challenge~\cite{3dpw_challenge}. Given the estimated SMPL pose $\hat{\pose}$, shape $\hat{\shape}$ and 3D joints $\hat{\mat{J}}$, corresponding GT pose $\pose$, shape $\shape$ and 3D joints $\mat{J}$, the evaluation metrics are defined as:

\begin{itemize}
    \item \textbf{MPJPE}: $e=\frac{1}{|\mat{J}|}\sum_i^{|\mat{J}|}||\mat{J}_i-\hat{\mat{J}}_i||_2$, Mean Per Joint Position Error (in mm). It measures the average Euclidean distance from prediction to ground truth joint positions. 
    \item \textbf{MPJPE-PA}: Mean Per Joint Position Error (in mm) after Procrustes analysis. (Rotation, translation, and scale are adjusted).
    \item \textbf{PCK}: percentage of correct joints. A joint is considered correct when it is less than 50mm away from the ground truth. 
    \item \textbf{AUC}: the total area under the PCK-threshold curve. It is calculated by computing PCKs by varying the error threshold from 0 to 200mm at which a predicted joint is considered correct. 
    \item \textbf{MPJAE}: $e=\frac{1}{K}\sum_k^{K}d_\text{geo}(\mat{R}_k, \hat{\mat{R}}_k)$, Mean Per Joint Angle Error (in mm), where $K$ is the total number of parts and $\mat{R}_k$ is the orientation of part $k$, $d_\text{geo}$ is the geodesic distance between two rotations. It measures the angle in degrees between the predicted part orientation and the ground truth orientation. The orientation difference is measured as the geodesic distance in SO(3). The 9 parts considered are: left/right upper arm, left/right lower arm, left/right upper leg, left/right lower leg, and root.
    \item \textbf{MPJAE-PA}: It measures the angle in degrees between the predicted part orientation and the ground truth orientation after rotating all predicted orientations by one global rotation matrix obtained from the Procrustes matching step. 
\end{itemize}

\paragraph{Baseline.} Similar to the object pose estimation track, we take the outputs from CHORE~\cite{xie22chore} as the baseline for the human reconstruction track. CHORE takes the human estimation results of FrankMocap~\cite{rong2020frankmocap} as input and further optimizes human and object poses based on network predictions. Its performance is shown in \cref{tab:human-recon}. It can be seen that the MPJPE-PA (55.6mm) is already quite close to the best-performing method in the 3DPW challenge (49.6mm), indicating that the existing human reconstruction method can already perform well on BEHAVE. 

\subsubsection{Joint reconstruction of human and object}
The two tracks discussed above reconstruct only humans or objects. However, the relative spatial arrangement of humans and objects during interaction is also important. Therefore, we propose the third track to reason both. 
\paragraph{Task definition.} Given a single image $\mat{I}\in \mathbb{R}^{3\times H\times W}$ of a person interacting with an object, the goal is to reconstruct the human and object in 3D. We represent humans using SMPL~\cite{smpl2015loper} and assume a known template mesh for the object. Therefore, the task boils down to estimating the SMPL pose $\pose\in \mathbb{R}^{3+72}$, shape $\shape\in \mathbb{R}^{10}$, and object rotation $\mat{R}_o\in \text{SO}(3)$ and translation $\vect{t}_o\in \mathbb{R}^3$ parameters. Following PHOSA~\cite{zhang2020phosa} and CHORE~\cite{xie22chore}, we provide human and object masks as optional inputs to the methods. 

\paragraph{Related work.} Interaction has been a popular research topic recently, with a focus on hand-object interaction from RGB~\cite{GrapingField:3DV:2020,Corona_2020_CVPR,hasson19_obman, ehsani2020force, yang2021cpf}, RGBD~\cite{Brahmbhatt_2019_CVPR,Brahmbhatt_2020_ECCV} or full 3D~\cite{GRAB:2020,ContactGrasp2019Brahmbhatt, zhou2022toch} input. Full-body interaction with dynamic large objects is less explored. PHOSA~\cite{zhang2020phosa} was the first attempt in this direction but it relies on manually defined interaction rules to optimize humans and objects, which is not scalable. CHORE~\cite{xie22chore} proposed a learning-based approach to learn interaction priors from data~\cite{bhatnagar22behave} and shows promising results. VisTracker~\cite{xie2023vistracker} further extends CHORE to deal with video input. Overall, research in this direction is still at a very early stage, calling for more innovative solutions for robust reconstruction. 

\paragraph{Evaluation.}
We adopt the metrics from CHORE~\cite{xie22chore} to compare different methods. Specifically, we first compute a Procrustes alignment on the combined SMPL and object vertices to align reconstruction with ground truth vertices. We then compute the SMPL and object error (unit: mm) of the reconstructed meshes as below:
\begin{itemize}
    \item SMPL: the Chamfer distance between reconstructed SMPL mesh and ground truth SMPL mesh.
    \item Object: the Chamfer distance between reconstructed object mesh and ground truth object mesh.
\end{itemize}
We sample 6k points from each mesh surface to compute the Chamfer distance. 

\paragraph{Baseline.} CHORE~\cite{xie22chore} was the best-performing method for jointly reconstructing humans and objects on the BEHAVE dataset. We hence use it as the baseline for this track. Its performance is summarized in \cref{tab:quant_cd}. It can be seen that the reconstruction errors of both humans (55.8mm) and objects (106.6mm) are still high, which calls for innovative methods to further improve the accuracy. 

\subsection{Results summary}
Our challenge was publicly released on February 1st, 2023 and the submission portal was closed on May 16th, 2023. The challenge attracted more than 100 participants with more than 300 submissions. The detailed participants are summarized below:
\begin{itemize}
    \item Object 6DoF: 32 participants with 90+ submissions. 
    \item Human reconstruction: 75 participants with 150+ submissions. 
    \item Joint reconstruction: 35 participants with 60+ submissions. 
\end{itemize}
It can be seen that human reconstruction is the most popular track, which also coincides with the amount of human mesh recovery publications in the literature. The number of participants for object 6DoF pose and joint reconstruction is similar while there are more submissions from the object-only track. This indicates that more people are becoming interested in modeling interaction despite the difficulty of this task.

In terms of the results, all winning methods outperform our baselines from prior SoTA methods. More specifically, the average recall of object 6DoF pose increases from 0.3 to 0.6 (\cref{tab:object-6dof}). The MPJPE-PA of human reconstruction decreases from 55.6 to 18.4 (\cref{tab:human-recon}). The SMPL and object error of joint reconstruction decreases from 55.8mm and 106.6mm to 45.1mm and 87.2mm respectively (\cref{tab:quant_cd}). It can be seen that the human error has been reduced to a very small level, while there is still a large space to further improve the object pose and joint reconstruction error. Some representative error comparisons can be found in \cref{sec:methods}.

\section{Methods}\label{sec:methods}
In this section, we discuss in more detail the winning methods and results of each track. Overall, we can see that estimating 2D-3D correspondence is an important part for all methods, in the form of either dense correspondence (object 6DoF pose) or keypoints (human and joint reconstruction). On the other hand, while direct regression (one-stage) can achieve the best results for separate human or object reconstruction, the joint reconstruction task still requires two stages (regression and then fitting) to obtain the best result. More technical details are described next. 

\subsection{Object 6DoF pose estimation}\label{subsec:object-6dof}
\begin{figure*}
  \centering
  \begin{subfigure}{0.7\linewidth}
    \includegraphics[width=1.0\linewidth]{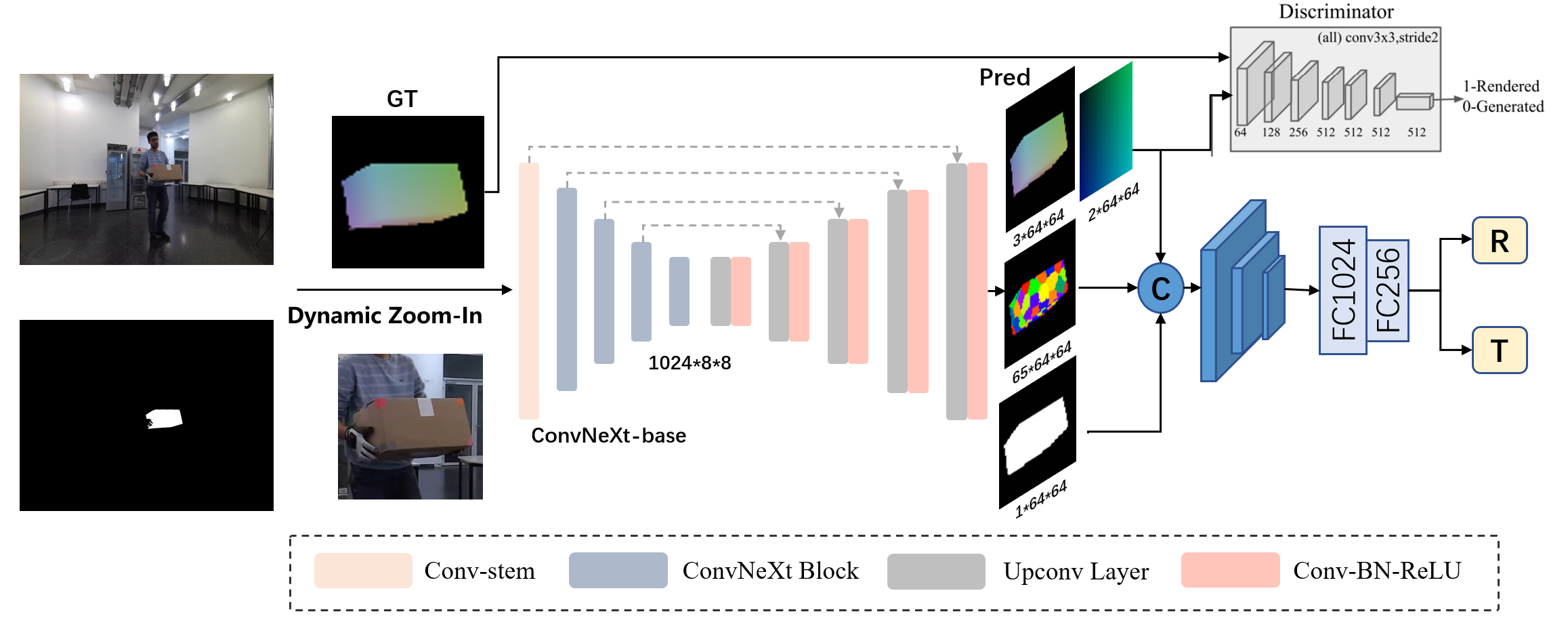}
    \caption{}
  \end{subfigure}
  \begin{subfigure}{0.25\linewidth}
   \includegraphics[width=0.8\linewidth]{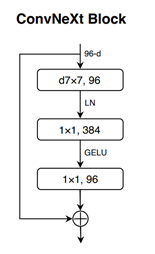}
    \caption{}
  \end{subfigure}
  \caption{ \textbf{Framework of G2DR} for object 6DoF pose estimation. Given an input image and a mask, we crop out the region of interest and predict intermediate features to directly regress the rotation and translation. Furthermore, we employ a generator-discriminator network to facilitate the generation of robust results in the presence of heavy occlusions.}
  \label{fig:object6dof-method}
\end{figure*}

The winning method of this track is presented by Zhang \etal from the University of Science and Technology of China. It combines GDR-Net~\cite{Wang_2021_CVPR} based on Pix2Pose \cite{park2019pix2pose} with GAN training. An overview of the method can be found in \cref{fig:object6dof-method}.

\subsubsection{Method detail}
\paragraph{Overview.} The BEHAVE dataset is built on various heavily occluded scenes where humans interact with objects, which imposes a great challenge on 6D object pose estimation. We hence leverage Pix2Pose~\cite{park2019pix2pose}, which could facilitate the training in heavily occluded occasions, together with GDR-Net~\cite{Wang_2021_CVPR}, which adopts the surface region attention module to mediate the influence of symmetric objects and regresses the 6D pose in an end-to-end manner.

\Cref{fig:object6dof-method} depicts an overview of our proposed method. Following GDR-net, we leverage 2D-3D correspondences represented as normalized object coordinate space (NOCS)~\cite{Wang_2019_CVPR_NOCS} to directly regress the 6D pose of objects. We further employ adversarial loss for training an effective NOCS map prediction network. To enhance the precision of our results, we pretrain our network with synthetic renderings of the BEHAVE objects. %

\paragraph{Object pose parametrization.} We use the popular 6D representation~\cite{Zhou_2019_CVPR} to represent object rotation, which has proven promising. For the translation, we utilize a Scale-Invariant representation for Translation Estimation (SITE) to deal with zoomed-in regions of interest (RoIs), same as GDR-Net~\cite{Wang_2021_GDRN}. %

\begin{figure}[h]
   \centering
   \includegraphics[width=1\linewidth]{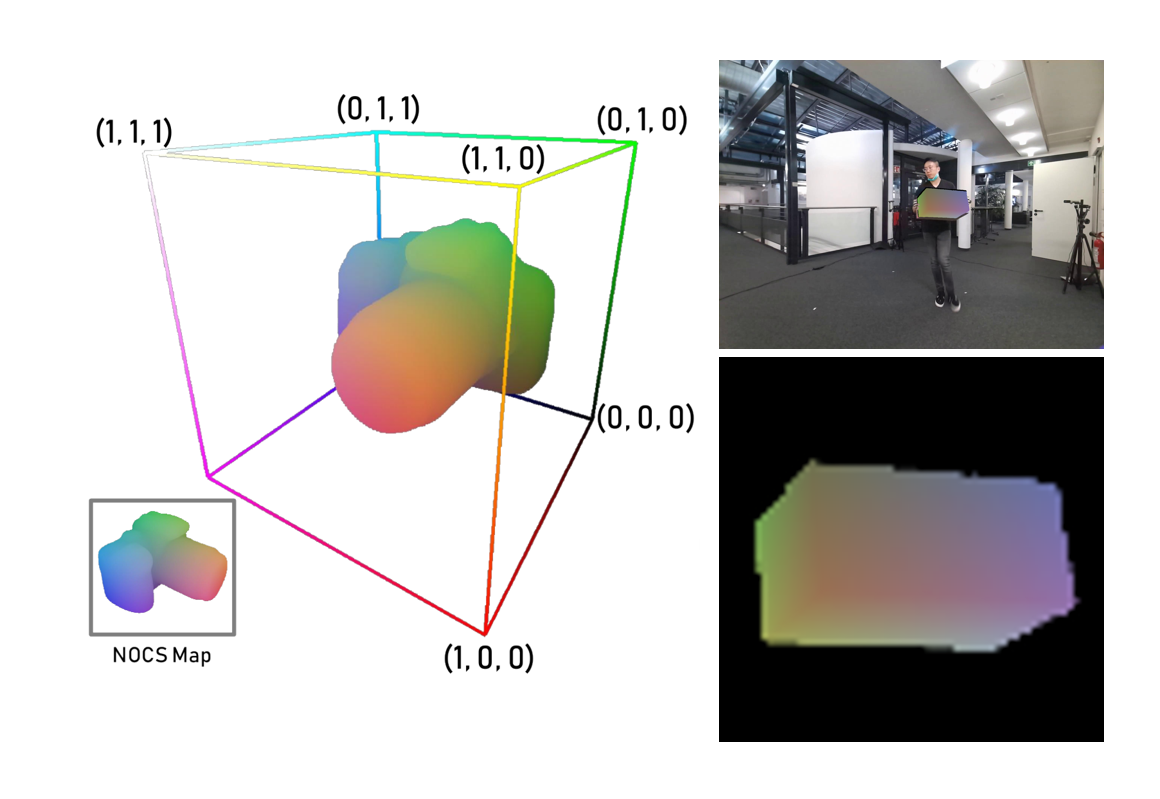}
    \caption{\textbf{Visualization of the 2D-3D correspondence} defined using the Normalized Object Coordinate Space (NOCS). The object model is normalized inside a unit cube. We then render the model as a 2D image where the color is the corresponding 3D coordinate.}
    \label{fig:nocs}
\end{figure}

\paragraph{Geometry-guided direct regression network (GDR-Net).} Our main block follows the same structure as GDR-Net~\cite{Wang_2021_GDRN}, see \cref{fig:object6dof-method}. During the training process, we predict a coarse-grained NOCS map named $M_{SRA}$~\cite{Wang_2021_CVPR} and dense per-pixel NOCS map $M_\text{2D-3D}$. A visualization of the NOCS map can be found in \cref{fig:nocs}. We follow GDR-Net~\cite{Wang_2021_GDRN} to compute a cross entropy (CE) loss $L_\text{CE}$ between the predicted coarse NOCS map $\hat{M_{SRA}}$ and GT map $M_{SRA}$. For the dense NOCS map $M_\text{2D-3D}$, GDR-Net computes an L1 distance between predicted and GT maps using a visibility mask $M_\text{vis}$. We additionally take the symmetry into account and compute the minimum loss of all symmetric rotations.

\paragraph{Training with GAN.} Attributed to the adversarial construction, Generative Adversarial Networks (GAN) have the capacity to discern and grasp the implicit information and distribution of the data. This characteristic empowers GAN to generate to realistic images in the target domain using images of other domains and facilitates the network to learn how to predict the occluded part based on the visible parts of the image.

The task for 3d coordinate regression is similar to this task since it seeks an implicit mapping between an RGB color and a 3D coordinate of an object. Therefore, the discriminator and the loss function $ \mathcal{L}_{GAN}$ can be employed to train the regression network. The discriminator network attempts to distinguish whether the 3D coordinate image is rendered using the gound-truth 3D model or is generated by our network. The loss is defined as,
\begin{equation}
    \mathcal{L}_{GAN} = log D(I_{gt}) + log(1 - D(G(I_{src}))),
\end{equation}
where $D$ denotes the discriminator network, and $I_{src}$ represents the NOCS map predicted by our network and $I_{gt}$ is the given ground truth. Finally, the objective of the training with GAN is formulated as

\begin{equation}
    G^{\star} = arg \min_{G} \max_{D} \mathcal{L}_{GAN}(G, D) + \lambda_{1}\mathcal{L}_{3D}(G),
\end{equation}
where $\lambda_{1}$ denotes weights to balance different tasks. 

\subsubsection{Training data preprocessing}
\par\textbf{Rendering of NOCS}: Firstly, we centralize given point cloud models to canonical models and normalize their sizes so that the boundaries of their max bounding box are within $[-1, 1]$. To render NOCS maps, we adopt BlenderProc based on discretized poses. The rendered color ranged in $[0,255]$ of each pixel in the 2D image corresponds to its 3D coordinate on the canonical model. 

\par\textbf{Synthetic datasets}: Due to the small size of the provided BEHAVE dataset and heavy occlusions in some cases, we employ data rendering techniques to simulate various occlusion scenarios and enhance the robustness and scalability of the model. We discretize the whole pose space represented by Euler angles into 8000 values and randomly initialize the translation. Using Blender software, we render each object to obtain RGB images, NOCS images, and accurate masks. To improve realism, we randomly place the rendered objects onto provided real background images. Furthermore, we randomly replace patches of the render figures with the background image to simulate the occluded occasions.

\par\textbf{Augmentation}: We have employed various data augmentation techniques to enhance the diversity and robustness of the image data. Firstly, we utilize the CoarseDropout technique, which applies coarse-grained occlusions to the image with a certain probability. Secondly, we apply Gaussian blur to the images by introducing random Gaussian kernels, resulting in a blurred effect that enhances the smoothness of the images. Additionally, we introduce random pixel-wise addition by randomly adding pixel values within a fixed range, allowing for random adjustments in the brightness of the images. Moreover, we introduce image inversion operations, which invert the colors of the images with a certain probability, thereby increasing the contrast and color diversity of the images. We also employ random pixel-wise multiplication by introducing random multiplication factors to randomly scale the pixel values of the images, enhancing the brightness and contrast of the images. Finally, we introduce contrast normalization, where random contrast scaling factors are applied to adjust the contrast of the images randomly, making the image details more prominent.

\par\textbf{2D detection}: Since we already have accurate mask images, we utilize these masks to locate the objects in RGB images. We use OpenCV to obtain the maximum bounding box of each object and employ an average strategy to determine the position of the center point. To include more global contexts, we enlarge the scale of the bounding box by 1.5 to capture more interactions between the object and the background. Additionally, to improve translation prediction, we introduce an offset which is a Gaussian distribution noise to the center point of the bounding box. This ensures that the area containing the object is approximately half the RoI.

\subsubsection{Implementation details}
All our experiments were conducted on a 24GB NVIDIA GeForce RTX 3090 GPU. During the training process, we set the batch size to 64 and utilize the Ranger optimizer. We employ a warm-up strategy to gradually increase the learning rate to 1e-5 and start cosine decay after 200 epochs. Due to the requirement that our network needs to converge on the prediction of NOCS maps before utilizing Patch-PnP to regress the correct pose, we set the coefficient of the $\mathcal{L}_{NOCS}$ loss to 10 in the first 200 epochs to prioritize the prediction of NOCS images. After 200 epochs, we increase the coefficient of the $\mathcal{L}_{Pose}$ loss in \cite{Wang_2021_CVPR}, used for direct pose regression, to 10 to facilitate the convergence of the network to accurate pose prediction.

\begin{figure*}
   \centering
   \begin{subfigure}{0.3\linewidth}
     \includegraphics[width=1.0\linewidth]{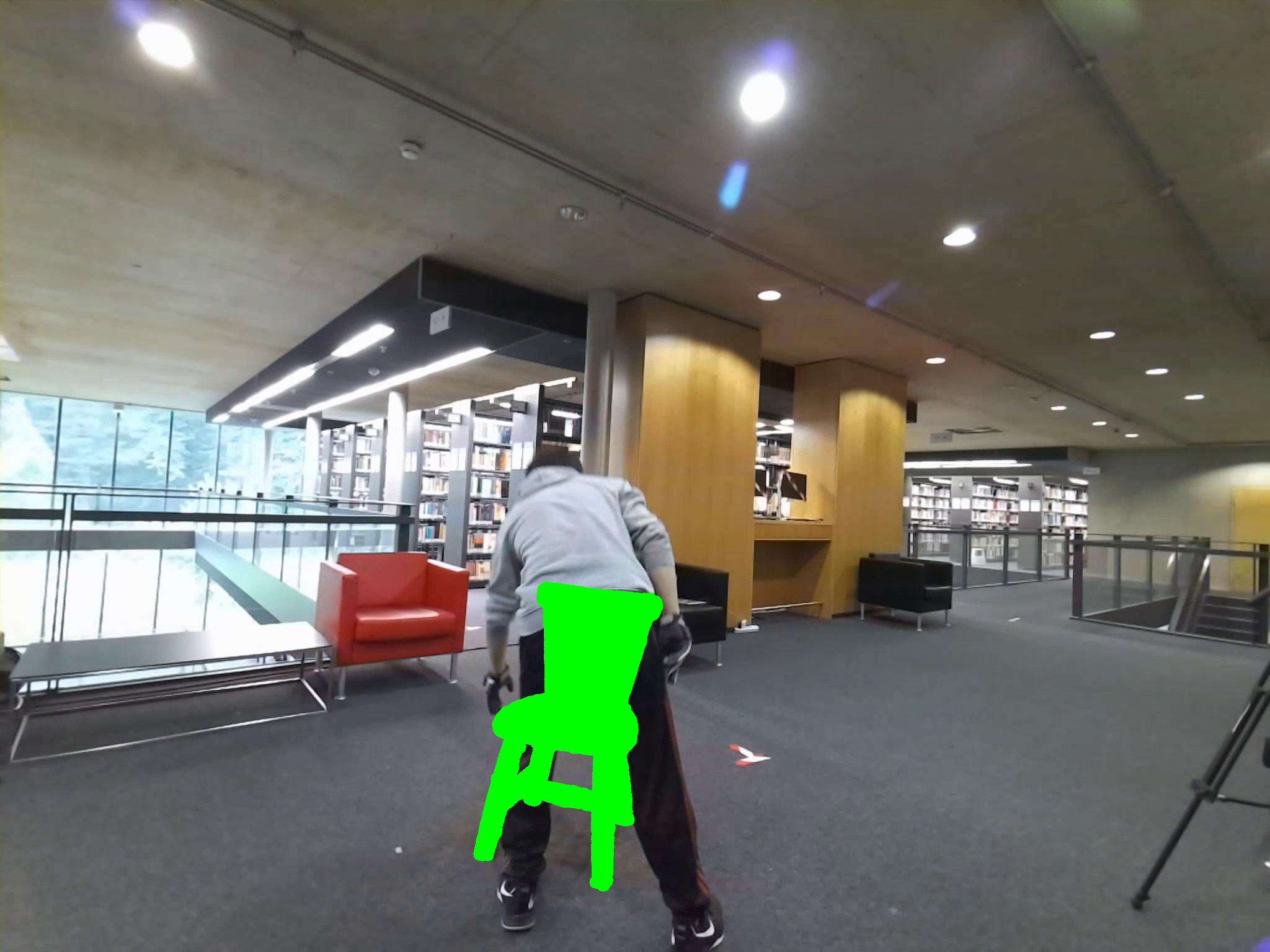}
     \caption{}
   \end{subfigure}
   \begin{subfigure}{0.3\linewidth}
    \includegraphics[width=1.0\linewidth]{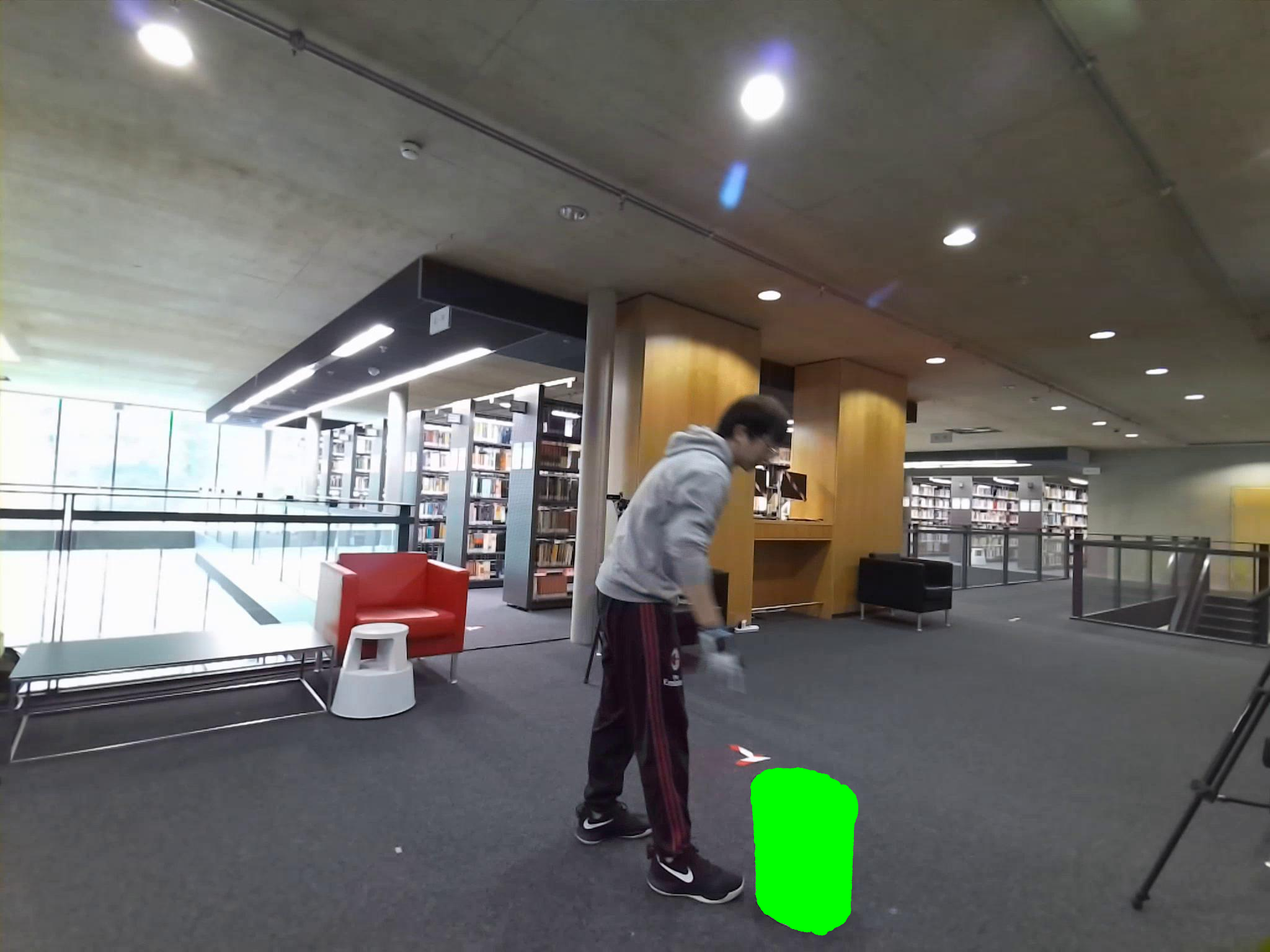}
     \caption{}
   \end{subfigure}
   \begin{subfigure}{0.3\linewidth}
    \includegraphics[width=1.0\linewidth]{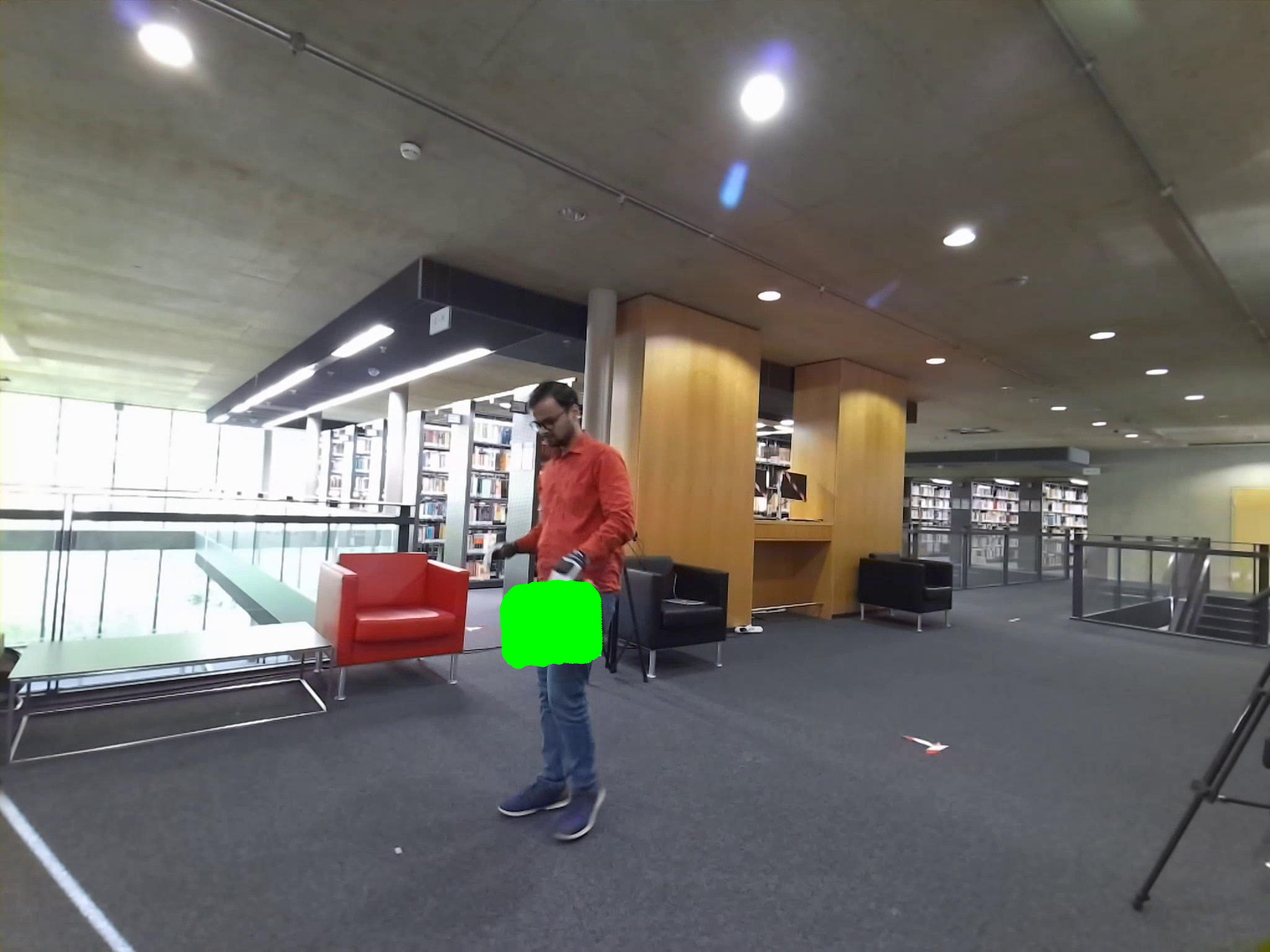}
     \caption{}
   \end{subfigure}
  \caption{ \textbf{Object 6DoF pose estimation on BEHAVE test dataset.} We present 3 examples of 6D pose estimation results. Given an input image, we show that the object model can be projected to the correct position based on the estimated pose and demonstrate our method performs well on strong occluded occasions. }
   \label{fig:result-6dof}
\end{figure*}

\subsubsection{Results}
 Under the setting above, Figure \ref{fig:result-6dof} reveals some examples of 6D pose estimation results. 
\begin{table*}[t]
    \centering
    \begin{tabular}{c|c c c c c c c}
       Method  & MSSD $\downarrow$ & MSPD $\downarrow$ & RE $\downarrow$ & MSSD-AR $\uparrow$ &MSPD-AR  $\uparrow$ & RE-AR  $\uparrow$ & AR-all  $\uparrow$\\
       \hline
        CHORE~\cite{xie22chore} & 0.5 &	110.0 &	43.0 &	0.2 &	0.3 &	0.4 &	0.3 \\ 
        {\bf G2DR} & {\bf 0.2} &	{\bf 54.5} &	{\bf 21.2} &	{\bf 0.6} &	{\bf 0.6} &	{\bf 0.5} &	{\bf 0.6}  \\ 
    \end{tabular}
    \caption{\textbf{Quantitative comparison} with previous SoTA methods on the object 6DoF pose estimation task. 
    }
    \label{tab:object-6dof}
\end{table*}

\subsection{Human Reconstruction}\label{subsec:human-recon}
\begin{figure*}[t]
    \centering
    \includegraphics[width=\linewidth]{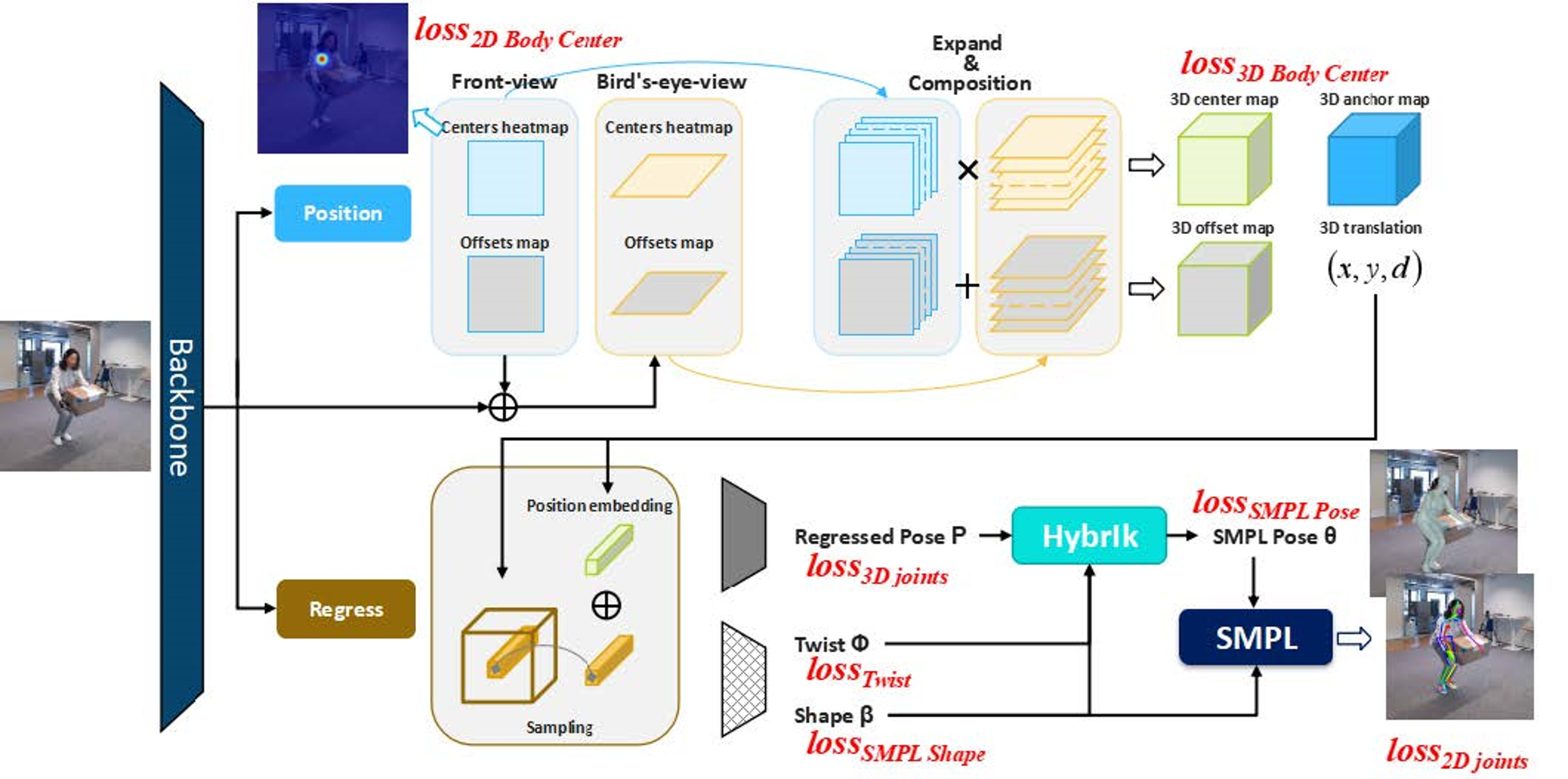}
    \caption{\textbf{Overview of the human reconstruction method}. We leverage BEV~\cite{Sun_CVPR2022_BEV} for global body center prediction and HybrIK~\cite{li2021hybrik} for accurate SMPL pose regression from predicted 3D joints. The network is trained end to end with losses highlighted in red. 
    }
    \label{fig:method-human-recon}
\end{figure*}
The winning method for the human reconstruction task is presented by Hao Chen and Xia Jia. The method combines BEV~\cite{Sun_CVPR2022_BEV} with HybrIK~\cite{li2021hybrik} and additional model ensemble tricks to achieve the best results. An overview of the method can be found in \cref{fig:method-human-recon}. 

\subsubsection{Method detail}
Given a single RGB image $\mat{I}\in \mathbb{R}^{3\times H \times W}$, we first use a backbone network to extract image features, which are sent to two branches to regress body center and joint locations respectively. We then use HybrIK to obtain SMPL pose parameters. 

\paragraph{The BEV branch} aims at predicting the human body center. From feature $\mat{F}\in \mathbb{R}^{D\times H^\prime \times W^\prime}$ extracted from input image, it regresses a front view center heatmap of size $\mathbb{R}^{1\times H\times W}$ and a bird's eye view heatmap of size $\mathbb{R}^{1\times D\times W}$. The second heatmap represents the likelihood of the person at some point in depth. Note that it does not represent metric depth. BEV then combines and refines these two maps into a 3D heatmap, $\mat{M}_{C}^{3D} \in \mathbb{R}^{1\times D\times H\times W}$. This heatmap is a representation of the human body center with 3D Gaussian Kernels. With this discretized center heatmap, we can only obtain a coarse prediction of the human location. We hence use an additional map that predicts an offset for each position. Formally, from the front-view and bird's eye view heatmap, we predict $\mat{M}_{O}^{3D} \in \mathbb{R}^{3\times D\times H\times W}$ to obtain more accurate position estimation. Please refer to \cite{Sun_CVPR2022_BEV} for more implementation details. 

\paragraph{Full body pose.} We then use another branch to estimate the body pose and shape parameters based on HybrIK~\cite{li2021hybrik}. We adopt the architecture from HybrIK to regress 3D body joints, twist angles, and SMPL shape parameters from image features and positional embedding of the 3D body center. These parameters are then sent to a differentiable inverse kinematic solver to obtain relative rotation (SMPL pose) parameters. The final mesh can be obtained via standard SMPL forward kinematics. See \cite{li2021hybrik} for more details. 

\paragraph{Training losses.} The method is trained end to end and the losses are highlighted in red in \cref{fig:method-human-recon}. The BEV branch is supervised with the focal loss~\cite{ROMP} formulated around 2D heatmap and 3D heatmap. The HybrIK branch is trained with L1 loss on the regressed 3D joints and L2 loss on the predicted twist angles and SMPL shape parameters. Additionally, we compute L2 loss on the finally estimated SMPL pose parameter and 2D joints after projection. 

\subsubsection{Additional tricks and results}
\begin{figure}
    \centering
    \includegraphics[width=\linewidth]{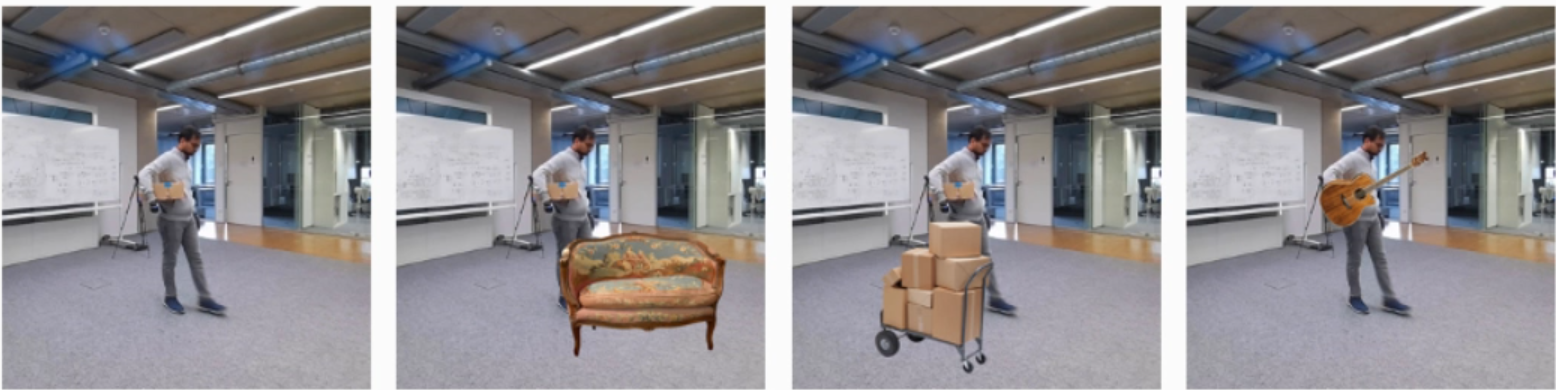}
    \caption{\textbf{Example data augmentation} for human reconstruction. We add obstacles to simulate occlusions which are typical in the BEHAVE dataset.}
    \label{fig:human-recon-augm}
\end{figure}
To enhance the performance of the model, we employed a series of tricks for optimization. These included but were not limited to, occlusion data augmentation, angle correction, multiscale training and testing, multi-stage training, and model ensemble. We use various data augmentations to enhance the training set to simulate the occlusions presented in BEHAVE. The most representative augmentation is done by introducing random obstacles and overlaps with the human body, some examples can be found in \cref{fig:human-recon-augm}. 

To further boost the performance, we adopt a stacking model ensemble technique to combine several models. Specifically, we train $M$ different models to predict the SMPL meshes. We then use the stacking ensemble technique to learn to combine these $M$ predictions into a single prediction. The learning is based on the validation set. We then transform the combined vertices to SMPL parameters using HybrIK to obtain the final SMPL predictions. 

We ablate the effectiveness of data augmentation and model ensemble in \cref{tab:human-recon}. It can be seen that both data augmentation and model ensemble contribute to the best results. 
\begin{table*}[]
    \centering
    \begin{tabular}{c|c c c c c c}
         Method & MPJPE $\downarrow$ & MPJPE-PA $\downarrow$ & PCK $\uparrow$ & AUC $\uparrow$ & MPJAE $\downarrow$ & MPJAE-PA $\downarrow$  \\
         \hline
         Baseline CHORE~\cite{xie22chore} & 100.8 &	55.6 &	26.6 &	0.5 &	19.6 &	15.9   \\ 
         BEV-HybrIK & 30.02 & 24.33 & 80.72 & 0.84 & 9.58 & 9.37 \\
         BEV-HybrIK (Data augm.) & 28.16 & 22.01 & 82.97 & 0.85 & 9.2 & 8.92 \\
         {\bf BEV-HybrIK (Data augm. + ensemble)} & {\bf 24.1} & {\bf 18.4} & {\bf 88.6} & {\bf 0.9} & {\bf 8.3} & {\bf 8.1} \\
    \end{tabular}
    \caption{\textbf{Ablating different tricks for human reconstruction}. It can be seen that both data augmentation and model ensemble improves the performance.}
    \label{tab:human-recon}
\end{table*}

\paragraph{Implementation.} We use HRNet64~\cite{Wang19HRNet} as the backbone network to extract image features and the input image resolution is 640. We ensemble $M=26$ models to obtain the final output.

\subsection{Joint human object reconstruction}\label{subsec:recon-hoi}
\begin{figure*}[t]
\begin{center}
\includegraphics[width=1.0\linewidth]{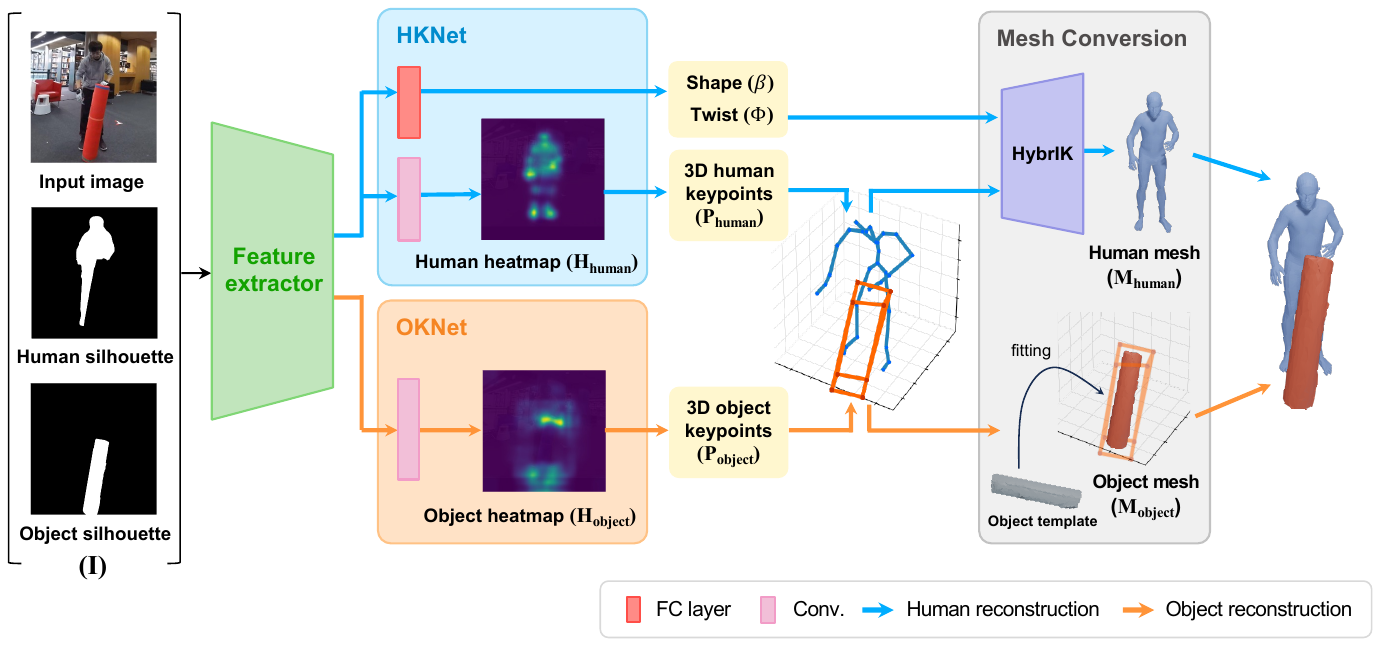}
\end{center}
\vspace*{-4.0mm}
\caption{\textbf{The overall pipeline of our proposed KHOR}, a keypoint-based 3D human and object reconstruction network. After extracting features using a backbone network, we use two networks to predict the heatmaps for human and object keypoints, which are then used to estimate the pose parameters of SMPL and object template meshes. 
}
\label{fig:model-hoi}
\vspace*{-2.0mm}
\end{figure*}

The winning method of the joint reconstruction track is presented by Nam \etal from Seoul National University. The method is called keypoint-based 3D human and object reconstruction (\textbf{KHOR}).
KHOR first estimates 3D keypoints of humans and objects and then converts the estimated 3D keypoints to 3D human and object meshes.
Figure~\ref{fig:model-hoi} shows an overall pipeline of the proposed method.

\subsubsection{Method details}
Our KHOR consists of four core components: feature extractor, HKNet, OKNet, and Mesh Conversion. These modules are detailed next.

\paragraph{Feature extractor.}
Given a concatenated input~of~image, human silhouette, and object silhouette~$\mathbf{I} \in \mathbb{R}^{(3 + 1 + 1) \times H \times W}$, a feature extractor extracts an image feature~$\mathbf{F} \in \mathbb{R}^{C \times H/4 \times W/4}$.
$H=256$, $W=256$, and $C=256$ represent image height, width, and channel dimension, respectively. 
We use HRNet-W48~\cite{sun2019deep} as the feature extractor after removing the last convolutional layer of the original HRNet-W48.

\paragraph{Human keypoint network~(HKNet).}
Human keypoint network~(HKNet) estimates 3D human keypoints~$\mathbf{P}_{\text{human}} \in \mathbb{R}^{29 \times 3}$ along with shape parameter~$\beta \in \mathbb{R}^{10}$ and twist angles~$\Phi \in \mathbb{R}^{46}$ from the extracted image feature~$\mathbf{F}$.
The 3D human keypoints $\mathbf{P}_{\text{human}}$ represent human joints (\textit{e.g.}, shoulder and wrist), defined in HybrIK~\cite{li2021hybrik}.
The shape parameter~$\beta$ represents PCA coefficients of T-posed human body shape space, and the twist angles~$\Phi$ denote rotation angles of human joints where each angle is defined with the corresponding bone as a rotation axis.
By forwarding the image feature~$\mathbf{F}$ into a 1-by-1 convolutional layer, a 3D heatmap~$\mathbf{H}_{\text{human}}$ is generated.
The 3D heatmap~$\mathbf{H}_{\text{human}}$ is used to regress 3D human keypoints~$\mathbf{P}_{\text{human}}$ via soft-argmax operation.
In addition, we use two separate fully-connected layers to predict shape parameter~$\beta$ and twist angles~$\Phi$ from the image feature~$\mathbf{F}$, respectively.

\paragraph{Object keypoint network~(OKNet).}
Object keypoint network (OKNet) directly estimates 3D object keypoints~$\mathbf{P}_{\text{object}} \in \mathbb{R}^{8 \times 3}$ from the extracted image feature~$\mathbf{F}$.
The object keypoints~$\mathbf{P}_{\text{object}}$ are defined by manually selected 3D corner points of the object mesh templates.
Like HKNet, 3D heatmap~$\mathbf{H}_{\text{object}}$ is generated by forwarding the image feature~$\mathbf{F}$ into a 1-by-1 convolutional layer.
We also regress the 3D object keypoints~$\mathbf{P}_{\text{object}}$ from the 3D heatmap~$\mathbf{H}_{\text{object}}$ with soft-argmax operation.

\paragraph{Mesh Conversion.}
In this module, the predicted 3D human keypoints~$\mathbf{P}_{\text{human}}$ from HKNet and 3D object keypoints~$\mathbf{P}_{\text{object}}$ from OKNet each are converted into 3D human mesh~$\mathbf{M}_{\text{human}}$ and 3D object mesh~$\mathbf{M}_{\text{object}}$.
To obtain human mesh~$\mathbf{M}_{\text{human}}$, we use HybrIK~\cite{li2021hybrik}, which solves SMPL pose parameter~$\theta \in \mathbb{R}^{72}$ from 3D human keypoints~$\mathbf{P}_{\text{human}}$, shape parameter~$\beta$, and twist angles~$\Phi$, via inverse kinematics process.
By passing the solved pose parameter~$\theta$ and shape parameter~$\beta$ into SMPL model~\cite{loper2015smpl}, we obtain 3D human mesh~$\mathbf{M}_{\text{human}}$.
To obtain an object mesh~$\mathbf{M}_{\text{object}}$, we position a 3D object template onto the proper 3D space by fitting its pre-defined 3D keypoints to predicted 3D object keypoints~$\mathbf{P}_{\text{object}}$.
Finally, we integrate the obtained 3D human mesh~$\mathbf{M}_{\text{human}}$ and 3D object mesh~$\mathbf{M}_{\text{object}}$ as our final reconstruction output.

\subsubsection{Implementation}
To train our KHOR, we minimize a loss function, defined as the combination of L2 distances between the predicted and ground truths~(GT) of~$P_{\text{human}}$,~$P_{\text{object}}$,~$\beta$, and~$\Phi$.
Additionally, following Li~\etal~\cite{li2021hybrik}, we include Laplace distribution loss in the final loss function to supervise the 3D human heatmap $\mathbf{H}_{\text{human}}$ and the 3D object heatmap $\mathbf{H}_{\text{object}}$.

For training and test stages, following previous work~\cite{xie22chore}, the concatenated input of image, human silhouette, and object silhouette are each cropped using a GT bounding box of both human and object.
Our feature extractor is initialized with publicly released pre-trained weights~\cite{sun2019deep} on ImageNet~\cite{deng2009imagenet}.
For training, the weights are updated by Adam~\cite{kingma2014adam} optimizer with a mini-batch size of 12.
Data augmentations, including scaling, rotation, and color jittering are performed during training. 
Our initial learning rate is set as~$1\times 10^{-3}$ and reduced by learning rate decay with a decay rate of 10 in the 60$^{\text{th}}$ and 80$^{\text{th}}$ epoch.
Our final model is trained for 90 epochs with a single NVIDIA RTX 2080 Ti GPU.

\subsubsection{Results}
We compare our method against baseline CHORE~\cite{xie22chore} and PHOSA~\cite{zhang2020phosa} in \cref{tab:quant_cd}. It can be seen that our method obtains better reconstruction results for both humans and objects. Moreover, our method runs much faster than baseline CHORE, which can be found in \cref{tab:quant_speed}. CHORE requires around 2000 network forward steps (NFE) to finish optimizing one frame, which costs more than 300 seconds, while our method only requires one forward step and takes only 0.115 seconds. 

\begin{table}[t]
\def\arraystretch{1.35}
\renewcommand{\tabcolsep}{0.7mm}
\begin{center}
\begin{tabular}{ >{\centering\arraybackslash}m{2.5cm}|>{\centering\arraybackslash}m{1.3cm}>{\centering\arraybackslash}m{1.3cm}}
\specialrule{.1em}{.05em}{0.0em}
  Methods & SMPL~$\downarrow$ & Object~$\downarrow$ \\ \hline
   PHOSA~\cite{zhang2020phosa} & 121.7 & 266.2 \\
   CHORE~\cite{xie22chore} & 55.8 & 106.6 \\
   \textbf{KHOR~(Ours)} & \textbf{45.1} & \textbf{87.2} \\
\specialrule{.1em}{-0.05em}{-0.05em}
\end{tabular}
\end{center}
\vspace*{-1.em}
\caption{\textbf{Quantitative comparison} with previous SOTA methods (joint reconstruction track). Our method obtains more accurate reconstruction for both humans and objects.}
\label{tab:quant_cd}
\end{table}

\begin{table}[t]
\def\arraystretch{1.35}
\renewcommand{\tabcolsep}{0.7mm}
\begin{center}
\begin{tabular}{>{\centering\arraybackslash}m{2.3cm}|>{\centering\arraybackslash}m{1.5cm}>
{\centering\arraybackslash}m{2.3cm}}
\specialrule{.1em}{.05em}{0.0em}
Methods & NFE $\downarrow$ & \makecell[c]{Running time $\downarrow$ \\ (sec / frame)}  \\ \hline
CHORE~\cite{xie22chore} & $\sim$ 2000 & 312.2 \\
\textbf{KHOR~(Ours)} & \textbf{1} & \textbf{0.115} \\
\specialrule{.1em}{-0.05em}{-0.05em}
\end{tabular}
\end{center}
\vspace*{-1.em}
\caption{\textbf{Speed comparison} with SoTA method CHORE~\cite{xie22chore}. CHORE requires evaluating the network forward function (NFE) many more times hence the speed is significantly slower than our method. 
}
\label{tab:quant_speed}
\end{table}

\section{Future Directions}
Our first RHOBIN challenge has pushed the field towards more accurate interaction capture from single RGB images. However, the research on interaction modelling and capture is still at a very early stage. There are still lots of research directions to be explored. We discuss some examples below. 

Firstly, current approaches focus on single images as input. In reality, interactions happen within a continuous timeslot and it is natural to capture interaction data as a sequence of images. Extending single frame methods to take video as input has been explored in human reconstruction \cite{VIBE:CVPR:2020, yuan2022glamr} and object pose estimation~\cite{wen_bundletrack_2021, bundlesdfwen2023}. They have shown that temporal information is useful for more consistent reconstruction and tracking. A recent work VisTracker~\cite{xie2023vistracker} also shows that temporal information helps obtain robust reconstruction even under heavy occlusion, which is not possible given a single image. However, it struggles with long-term occlusion and fast motion. Future works can further explore motion priors to assist video-based reconstruction and tracking. 

Secondly, current approaches assume a known category-level object template for the object to be reconstructed, which limits its applicability to general in the wild setting. Furthermore, the object pose estimator is trained specifically for the object template, which does not generalize well when new object geometry is different, e.g. different chairs. Template-free reconstruction of human \cite{pifuSHNMKL19, saito2020pifuhd} or object \cite{genre, i_DeepSDF, chibane20ifnet, DISN} has been studied separately yet the joint reconstruction remains unexplored due to the lack of data. A recent method~\cite{xie2023template_free} generates large synthetic data to learn both interaction and shape priors together. Future works can further explore this direction to obtain a more robust reconstruction. Another direction is to leverage temporal information to build the object template on the fly, which has been studied in hand-object interaction~\cite{hampali2023inhand, Ye_2023_ICCV, Rozumnyi_2023_ICCV} or general articulation objects without occlusion~\cite{yang2021lasr, yang2022banmo, yang2023rac}. To reconstruct the object, these methods~\cite{hampali2023inhand, Ye_2023_ICCV, Rozumnyi_2023_ICCV} leverage the constraint of hand articulation motion during hand-object interaction to build a template for the object. Full-body interaction is however more complex since the constraint on object shape imposed by the human body is less and the interaction is more complex. There is only one recent work~\cite{bundlesdfwen2023} that can handle this heavy occlusion using monocular RGBD videos. Modelling from monocular RGB video is still an interesting direction for further study. 

Beyond the shape and geometry, the human and object appearance remains an important part of a holistic understanding of the scene. BEHAVE~\cite{bhatnagar22behave} captures the 3D static scan of the human and object but does not capture the full 3D appearance information during the interaction. A more recent work~\cite{zhang2023neuraldome} captures the appearance as well as clothing deformation of the subject, which makes it possible to also evaluate methods that reconstruct appearance. Along this direction, recent work~\cite{kerbl3Dgaussians} proposes an efficient representation, Gaussian Splatting, that allows fast rendering at 100fps and flexible manipulation. Future works can further explore this representation for modelling interactions with appearance.  

Last but not least, the current challenge setting is limited to single-person interaction with one object. Real-life interaction can involve multi-person~\cite{Fieraru_2020_CVPR} and multi-objects~\cite{PROX:2019}, which is a more complex scenario which requires not only new benchmark datasets but also novel methods for representation learning and reconstruction. These are also interesting directions for future research.

\section{Conclusion}\label{sec:conclusion}
Human reconstruction and object pose estimation were mostly two separate fields in which researchers work independently. Our first RHOBIN challenge aimed at bringing two fields together to exchange intuitions and spark new ideas. The RHOBIN challenge consists of three tracks on human reconstruction, object 6DoF pose estimation and joint reconstruction. In this paper, we describe the evaluation and details about the winning methods of each track. All winning methods outperformed the baselines from previous SoTA publications. We also found that the performance of the human reconstruction method is mature, while there is still a large space to further improve object pose estimation and joint reconstruction. Future works can explore temporal information to obtain more robust or template-free reconstructions or extend interaction beyond a single object. We will continue to organize the RHOBIN challenge to encourage collaboration in this emerging research field.

{
    \small
    \bibliographystyle{ieeenat_fullname}
    \bibliography{main}

\begin{thebibliography}{78}
\providecommand{\natexlab}[1]{#1}
\providecommand{\url}[1]{\texttt{#1}}
\expandafter\ifx\csname urlstyle\endcsname\relax
  \providecommand{\doi}[1]{doi: #1}\else
  \providecommand{\doi}{doi: \begingroup \urlstyle{rm}\Url}\fi

\bibitem[3dp()]{3dpw_challenge}
https://virtualhumans.mpi-inf.mpg.de/3dpw\_challenge/.

\bibitem[rho()]{rhobin2023}
https://rhobin-challenge.github.io/.

\bibitem[Bhatnagar et~al.(2022)Bhatnagar, Xie, Petrov, Sminchisescu, Theobalt, and Pons-Moll]{bhatnagar22behave}
Bharat~Lal Bhatnagar, Xianghui Xie, Ilya Petrov, Cristian Sminchisescu, Christian Theobalt, and Gerard Pons-Moll.
\newblock Behave: Dataset and method for tracking human object interactions.
\newblock In \emph{{IEEE} Conference on Computer Vision and Pattern Recognition (CVPR)}, 2022.

\bibitem[Bogo et~al.(2016)Bogo, Kanazawa, Lassner, Gehler, Romero, and Black]{bogo2016smplify}
Federica Bogo, Angjoo Kanazawa, Christoph Lassner, Peter Gehler, Javier Romero, and Michael~J Black.
\newblock Keep it {SMPL}: Automatic estimation of {3D} human pose and shape from a single image.
\newblock In \emph{European Conf. on Computer Vision}. Springer International Publishing, 2016.

\bibitem[Brahmbhatt et~al.(2019{\natexlab{a}})Brahmbhatt, Ham, Kemp, and Hays]{Brahmbhatt_2019_CVPR}
Samarth Brahmbhatt, Cusuh Ham, Charles~C. Kemp, and James Hays.
\newblock {ContactDB}: Analyzing and predicting grasp contact via thermal imaging.
\newblock In \emph{The IEEE Conference on Computer Vision and Pattern Recognition (CVPR)}, 2019{\natexlab{a}}.

\bibitem[Brahmbhatt et~al.(2019{\natexlab{b}})Brahmbhatt, Handa, Hays, and Fox]{ContactGrasp2019Brahmbhatt}
Samarth Brahmbhatt, Ankur Handa, James Hays, and Dieter Fox.
\newblock Contactgrasp: Functional multi-finger grasp synthesis from contact.
\newblock In \emph{IROS}, 2019{\natexlab{b}}.

\bibitem[Brahmbhatt et~al.(2020)Brahmbhatt, Tang, Twigg, Kemp, and Hays]{Brahmbhatt_2020_ECCV}
Samarth Brahmbhatt, Chengcheng Tang, Christopher~D. Twigg, Charles~C. Kemp, and James Hays.
\newblock {ContactPose}: A dataset of grasps with object contact and hand pose.
\newblock In \emph{The European Conference on Computer Vision (ECCV)}, 2020.

\bibitem[Chibane et~al.(2020)Chibane, Alldieck, and Pons-Moll]{chibane20ifnet}
Julian Chibane, Thiemo Alldieck, and Gerard Pons-Moll.
\newblock Implicit functions in feature space for 3d shape reconstruction and completion.
\newblock In \emph{{IEEE} Conference on Computer Vision and Pattern Recognition (CVPR)}. {IEEE}, 2020.

\bibitem[Corona et~al.(2020)Corona, Pumarola, Alenya, Moreno-Noguer, and Rogez]{Corona_2020_CVPR}
Enric Corona, Albert Pumarola, Guillem Alenya, Francesc Moreno-Noguer, and Gregory Rogez.
\newblock Ganhand: Predicting human grasp affordances in multi-object scenes.
\newblock In \emph{Proceedings of the IEEE/CVF Conference on Computer Vision and Pattern Recognition (CVPR)}, 2020.

\bibitem[Deng et~al.(2009)Deng, Dong, Socher, Li, Li, and Fei-Fei]{deng2009imagenet}
Jia Deng, Wei Dong, Richard Socher, Li-Jia Li, Kai Li, and Li Fei-Fei.
\newblock {ImageNet}: A large-scale hierarchical image database.
\newblock In \emph{CVPR}, 2009.

\bibitem[Di et~al.(2021)Di, Manhardt, Wang, Ji, Navab, and Tombari]{di_so-pose_2021}
Yan Di, Fabian Manhardt, Gu Wang, Xiangyang Ji, Nassir Navab, and Federico Tombari.
\newblock {SO}-{Pose}: {Exploiting} {Self}-{Occlusion} for {Direct} {6D} {Pose} {Estimation}.
\newblock In \emph{2021 {IEEE}/{CVF} {International} {Conference} on {Computer} {Vision} ({ICCV})}, pages 12376--12385, Montreal, QC, Canada, 2021. IEEE.

\bibitem[Ehsani et~al.(2020)Ehsani, Tulsiani, Gupta, Farhadi, and Gupta]{ehsani2020force}
Kiana Ehsani, Shubham Tulsiani, Saurabh Gupta, Ali Farhadi, and Abhinav Gupta.
\newblock Use the force, luke! learning to predict physical forces by simulating effects.
\newblock In \emph{CVPR}, 2020.

\bibitem[Fieraru et~al.(2020)Fieraru, Zanfir, Oneata, Popa, Olaru, and Sminchisescu]{Fieraru_2020_CVPR}
Mihai Fieraru, Mihai Zanfir, Elisabeta Oneata, Alin-Ionut Popa, Vlad Olaru, and Cristian Sminchisescu.
\newblock Three-dimensional reconstruction of human interactions.
\newblock In \emph{Proceedings of the IEEE/CVF Conference on Computer Vision and Pattern Recognition (CVPR)}, 2020.

\bibitem[Guzov et~al.(2021)Guzov, Mir, Sattler, and Pons-Moll]{CVPR21HPS}
Vladimir Guzov, Aymen Mir, Torsten Sattler, and Gerard Pons-Moll.
\newblock Human poseitioning system (hps): 3d human pose estimation and self-localization in large scenes from body-mounted sensors.
\newblock In \emph{{IEEE} Conference on Computer Vision and Pattern Recognition (CVPR)}. {IEEE}, 2021.

\bibitem[Hampali et~al.(2023)Hampali, Hodan, Tran, Ma, Keskin, and Lepetit]{hampali2023inhand}
Shreyas Hampali, Tomas Hodan, Luan Tran, Lingni Ma, Cem Keskin, and Vincent Lepetit.
\newblock In-hand 3d object scanning from an rgb sequence.
\newblock \emph{CVPR}, 2023.

\bibitem[Hassan et~al.(2019)Hassan, Choutas, Tzionas, and Black]{PROX:2019}
Mohamed Hassan, Vasileios Choutas, Dimitrios Tzionas, and Michael~J. Black.
\newblock Resolving 3d human pose ambiguities with 3d scene constraints.
\newblock In \emph{International Conference on Computer Vision}, 2019.

\bibitem[Hasson et~al.(2019)Hasson, Varol, Tzionas, Kalevatykh, Black, Laptev, and Schmid]{hasson19_obman}
Yana Hasson, G{\"u}l Varol, Dimitrios Tzionas, Igor Kalevatykh, Michael~J. Black, Ivan Laptev, and Cordelia Schmid.
\newblock Learning joint reconstruction of hands and manipulated objects.
\newblock In \emph{CVPR}, 2019.

\bibitem[Hinterstoisser et~al.(2013)Hinterstoisser, Lepetit, Ilic, Holzer, Bradski, Konolige, and Navab]{linemod}
Stefan Hinterstoisser, Vincent Lepetit, Slobodan Ilic, Stefan Holzer, Gary Bradski, Kurt Konolige, and Nassir Navab.
\newblock Model based training, detection and pose estimation of texture-less 3d objects in heavily cluttered scenes.
\newblock In \emph{Computer Vision -- ACCV 2012}, pages 548--562, Berlin, Heidelberg, 2013. Springer Berlin Heidelberg.

\bibitem[Hoda{\v{n}} et~al.(2018)Hoda{\v{n}}, Michel, Brachmann, Kehl, Glent~Buch, Kraft, Drost, Vidal, Ihrke, Zabulis, Sahin, Manhardt, Tombari, Kim, Matas, and Rother]{hodan2018bop}
Tom{\'a}{\v{s}} Hoda{\v{n}}, Frank Michel, Eric Brachmann, Wadim Kehl, Anders Glent~Buch, Dirk Kraft, Bertram Drost, Joel Vidal, Stephan Ihrke, Xenophon Zabulis, Caner Sahin, Fabian Manhardt, Federico Tombari, Tae-Kyun Kim, Ji{\v{r}}{\'i} Matas, and Carsten Rother.
\newblock {BOP}: Benchmark for {6D} object pose estimation.
\newblock \emph{European Conference on Computer Vision (ECCV)}, 2018.

\bibitem[Hu et~al.(2020)Hu, Fua, Wang, and Salzmann]{hu_single-stage_cvpr20}
Yinlin Hu, Pascal Fua, Wei Wang, and Mathieu Salzmann.
\newblock Single-stage 6d object pose estimation.
\newblock In \emph{CVPR}, 2020.

\bibitem[Huang et~al.(2022)Huang, Taheri, Black, and Tzionas]{huang2022intercap}
Yinghao Huang, Omid Taheri, Michael~J. Black, and Dimitrios Tzionas.
\newblock {InterCap}: {J}oint markerless {3D} tracking of humans and objects in interaction.
\newblock In \emph{{German Conference on Pattern Recognition (GCPR)}}, pages 281--299. Springer, 2022.

\bibitem[Huang et~al.(2024)Huang, Yi, Xiu, Liao, Tang, Cai, and Thies]{huang2024tech}
Yangyi Huang, Hongwei Yi, Yuliang Xiu, Tingting Liao, Jiaxiang Tang, Deng Cai, and Justus Thies.
\newblock {TeCH: Text-guided Reconstruction of Lifelike Clothed Humans}.
\newblock In \emph{International Conference on 3D Vision (3DV)}, 2024.

\bibitem[Kanazawa et~al.(2018)Kanazawa, Black, Jacobs, and Malik]{hmrKanazawa17}
Angjoo Kanazawa, Michael~J. Black, David~W. Jacobs, and Jitendra Malik.
\newblock End-to-end recovery of human shape and pose.
\newblock In \emph{Computer Vision and Pattern Recognition (CVPR)}, 2018.

\bibitem[Karunratanakul et~al.(2020)Karunratanakul, Yang, Zhang, Black, Muandet, and Tang]{GrapingField:3DV:2020}
Korrawe Karunratanakul, Jinlong Yang, Yan Zhang, Michael Black, Krikamol Muandet, and Siyu Tang.
\newblock Grasping field: Learning implicit representations for human grasps.
\newblock In \emph{8th International Conference on 3D Vision}, pages 333--344. {IEEE}, 2020.

\bibitem[Kerbl et~al.(2023)Kerbl, Kopanas, Leimk{\"u}hler, and Drettakis]{kerbl3Dgaussians}
Bernhard Kerbl, Georgios Kopanas, Thomas Leimk{\"u}hler, and George Drettakis.
\newblock 3d gaussian splatting for real-time radiance field rendering.
\newblock \emph{ACM Transactions on Graphics}, 42\penalty0 (4), 2023.

\bibitem[Kingma and Ba(2014)]{kingma2014adam}
Diederik~P Kingma and Jimmy Ba.
\newblock Adam: A method for stochastic optimization.
\newblock In \emph{ICLR}, 2014.

\bibitem[Kocabas et~al.(2020)Kocabas, Athanasiou, and Black]{VIBE:CVPR:2020}
Muhammed Kocabas, Nikos Athanasiou, and Michael~J. Black.
\newblock {VIBE}: Video inference for human body pose and shape estimation.
\newblock In \emph{Proceedings IEEE Conf. on Computer Vision and Pattern Recognition (CVPR)}, pages 5252--5262. IEEE, 2020.

\bibitem[Kolotouros et~al.(2019)Kolotouros, Pavlakos, Black, and Daniilidis]{kolotouros2019spin}
Nikos Kolotouros, Georgios Pavlakos, Michael~J Black, and Kostas Daniilidis.
\newblock Learning to reconstruct 3d human pose and shape via model-fitting in the loop.
\newblock In \emph{ICCV}, 2019.

\bibitem[Li et~al.(2021)Li, Xu, Chen, Bian, Yang, and Lu]{li2021hybrik}
Jiefeng Li, Chao Xu, Zhicun Chen, Siyuan Bian, Lixin Yang, and Cewu Lu.
\newblock Hybrik: A hybrid analytical-neural inverse kinematics solution for 3d human pose and shape estimation.
\newblock In \emph{Proceedings of the IEEE/CVF Conference on Computer Vision and Pattern Recognition}, pages 3383--3393, 2021.

\bibitem[Li et~al.(2020)Li, Xiu, Saito, Huang, Olszewski, and Li]{li2020monocap}
Ruilong Li, Yuliang Xiu, Shunsuke Saito, Zeng Huang, Kyle Olszewski, and Hao Li.
\newblock Monocular real-time volumetric performance capture.
\newblock \emph{arXiv preprint arXiv:2007.13988}, 2020.

\bibitem[Li et~al.(2018)Li, Wang, Ji, Xiang, and Fox]{li_deepim_nodate}
Yi Li, Gu Wang, Xiangyang Ji, Yu Xiang, and Dieter Fox.
\newblock Deepim: Deep iterative matching for 6d pose estimation.
\newblock In \emph{European Conference on Computer Vision (ECCV)}, 2018.

\bibitem[Liao et~al.(2024)Liao, Yi, Xiu, Tang, Huang, Thies, and Black]{liao2023tada}
Tingting Liao, Hongwei Yi, Yuliang Xiu, Jiaxiang Tang, Yangyi Huang, Justus Thies, and Michael~J. Black.
\newblock {TADA! Text to Animatable Digital Avatars}.
\newblock In \emph{International Conference on 3D Vision (3DV)}, 2024.

\bibitem[Liu et~al.(2022)Liu, Wen, Peng, Lin, Long, Komura, and Wang]{liu2022gen6d}
Yuan Liu, Yilin Wen, Sida Peng, Cheng Lin, Xiaoxiao Long, Taku Komura, and Wenping Wang.
\newblock Gen6d: Generalizable model-free 6-dof object pose estimation from rgb images.
\newblock In \emph{ECCV}, 2022.

\bibitem[Loper et~al.(2015{\natexlab{a}})Loper, Mahmood, Romero, Pons-Moll, and Black]{loper2015smpl}
Matthew Loper, Naureen Mahmood, Javier Romero, Gerard Pons-Moll, and Michael~J Black.
\newblock {SMPL}: A skinned multi-person linear model.
\newblock In \emph{ACM TOG}, 2015{\natexlab{a}}.

\bibitem[Loper et~al.(2015{\natexlab{b}})Loper, Mahmood, Romero, Pons-Moll, and Black]{smpl2015loper}
Matthew Loper, Naureen Mahmood, Javier Romero, Gerard Pons-Moll, and Michael~J. Black.
\newblock {SMPL}: A skinned multi-person linear model.
\newblock In \emph{ACM Transactions on Graphics}. ACM, 2015{\natexlab{b}}.

\bibitem[Majcher and Kwolek(2022)]{majcher_shape_nodate}
Mateusz Majcher and Bogdan Kwolek.
\newblock Shape enhanced keypoints learning with geometric prior for 6d object pose tracking.
\newblock In \emph{2022 IEEE/CVF Conference on Computer Vision and Pattern Recognition Workshops (CVPRW)}, pages 2985--2991, 2022.

\bibitem[Park et~al.(2019{\natexlab{a}})Park, Florence, Straub, Newcombe, and Lovegrove]{i_DeepSDF}
Jeong~Joon Park, Peter Florence, Julian Straub, Richard~A. Newcombe, and Steven Lovegrove.
\newblock Deepsdf: Learning continuous signed distance functions for shape representation.
\newblock In \emph{{IEEE} Conference on Computer Vision and Pattern Recognition, {CVPR} 2019, Long Beach, CA, USA, June 16-20, 2019}, pages 165--174, 2019{\natexlab{a}}.

\bibitem[Park et~al.(2019{\natexlab{b}})Park, Patten, and Vincze]{park2019pix2pose}
Kiru Park, Timothy Patten, and Markus Vincze.
\newblock Pix2pose: Pixel-wise coordinate regression of objects for 6d pose estimation.
\newblock In \emph{Proceedings of the IEEE/CVF International Conference on Computer Vision}, pages 7668--7677, 2019{\natexlab{b}}.

\bibitem[Pavlakos et~al.(2019)Pavlakos, Choutas, Ghorbani, Bolkart, Osman, Tzionas, and Black]{SMPL-X:2019}
Georgios Pavlakos, Vasileios Choutas, Nima Ghorbani, Timo Bolkart, Ahmed A.~A. Osman, Dimitrios Tzionas, and Michael~J. Black.
\newblock Expressive body capture: 3d hands, face, and body from a single image.
\newblock In \emph{Proceedings IEEE Conf. on Computer Vision and Pattern Recognition (CVPR)}, 2019.

\bibitem[Peng et~al.(2019)Peng, Liu, Huang, Zhou, and Bao]{peng_pvnet_2019}
Sida Peng, Yuan Liu, Qixing Huang, Xiaowei Zhou, and Hujun Bao.
\newblock {PVNet}: {Pixel}-{Wise} {Voting} {Network} for {6DoF} {Pose} {Estimation}.
\newblock In \emph{2019 {IEEE}/{CVF} {Conference} on {Computer} {Vision} and {Pattern} {Recognition} ({CVPR})}, pages 4556--4565, Long Beach, CA, USA, 2019. IEEE.

\bibitem[Rong et~al.(2021)Rong, Shiratori, and Joo]{rong2020frankmocap}
Yu Rong, Takaaki Shiratori, and Hanbyul Joo.
\newblock Frankmocap: A monocular 3d whole-body pose estimation system via regression and integration.
\newblock In \emph{IEEE International Conference on Computer Vision Workshops}, 2021.

\bibitem[Rozumnyi et~al.(2023)Rozumnyi, Matas, Pollefeys, Ferrari, and Oswald]{Rozumnyi_2023_ICCV}
Denys Rozumnyi, Ji\v{r}{\'\i} Matas, Marc Pollefeys, Vittorio Ferrari, and Martin~R. Oswald.
\newblock Tracking by 3d model estimation of unknown objects in videos.
\newblock In \emph{Proceedings of the IEEE/CVF International Conference on Computer Vision (ICCV)}, pages 14086--14096, 2023.

\bibitem[Saito et~al.(2019)Saito, , Huang, Natsume, Morishima, Kanazawa, and Li]{pifuSHNMKL19}
Shunsuke Saito, , Zeng Huang, Ryota Natsume, Shigeo Morishima, Angjoo Kanazawa, and Hao Li.
\newblock Pifu: Pixel-aligned implicit function for high-resolution clothed human digitization.
\newblock In \emph{{IEEE} International Conference on Computer Vision ({ICCV})}. {IEEE}, 2019.

\bibitem[Saito et~al.(2020)Saito, Simon, Saragih, and Joo]{saito2020pifuhd}
Shunsuke Saito, Tomas Simon, Jason Saragih, and Hanbyul Joo.
\newblock Pifuhd: Multi-level pixel-aligned implicit function for high-resolution 3d human digitization.
\newblock In \emph{Proceedings of the IEEE Conference on Computer Vision and Pattern Recognition}, 2020.

\bibitem[Sun et~al.(2022{\natexlab{a}})Sun, Wang, Zhang, He, Zhao, Zhang, and Zhou]{sun2022onepose}
Jiaming Sun, Zihao Wang, Siyu Zhang, Xingyi He, Hongcheng Zhao, Guofeng Zhang, and Xiaowei Zhou.
\newblock {OnePose}: One-shot object pose estimation without {CAD} models.
\newblock \emph{CVPR}, 2022{\natexlab{a}}.

\bibitem[Sun et~al.(2019)Sun, Xiao, Liu, and Wang]{sun2019deep}
Ke Sun, Bin Xiao, Dong Liu, and Jingdong Wang.
\newblock Deep high-resolution representation learning for human pose estimation.
\newblock In \emph{CVPR}, 2019.

\bibitem[Sun et~al.(2021)Sun, Bao, Liu, Fu, Michael~J., and Mei]{ROMP}
Yu Sun, Qian Bao, Wu Liu, Yili Fu, Black Michael~J., and Tao Mei.
\newblock {Monocular, One-stage, Regression of Multiple 3D People}.
\newblock In \emph{ICCV}, 2021.

\bibitem[Sun et~al.(2022{\natexlab{b}})Sun, Liu, Bao, Fu, Mei, and Black]{Sun_CVPR2022_BEV}
Yu Sun, Wu Liu, Qian Bao, Yili Fu, Tao Mei, and Michael~J. Black.
\newblock Putting people in their place: Monocular regression of 3d people in depth.
\newblock In \emph{Proceedings of the IEEE/CVF Conference on Computer Vision and Pattern Recognition (CVPR)}, pages 13243--13252, 2022{\natexlab{b}}.

\bibitem[Taheri et~al.(2020)Taheri, Ghorbani, Black, and Tzionas]{GRAB:2020}
Omid Taheri, Nima Ghorbani, Michael~J. Black, and Dimitrios Tzionas.
\newblock {GRAB}: A dataset of whole-body human grasping of objects.
\newblock In \emph{European Conference on Computer Vision (ECCV)}, 2020.

\bibitem[Tian et~al.(2022)Tian, Zhang, Liu, and Wang]{tian2022hmrsurvey}
Yating Tian, Hongwen Zhang, Yebin Liu, and Limin Wang.
\newblock Recovering 3d human mesh from monocular images: A survey.
\newblock \emph{arXiv preprint arXiv:2203.01923}, 2022.

\bibitem[Tzionas et~al.(2016)Tzionas, Ballan, Srikantha, Aponte, Pollefeys, and Gall]{Tzionas2016capture-handobject}
Dimitrios Tzionas, Luca Ballan, Abhilash Srikantha, Pablo Aponte, Marc Pollefeys, and Juergen Gall.
\newblock Capturing hands in action using discriminative salient points and physics simulation.
\newblock \emph{International Journal of Computer Vision (IJCV)}, 2016.

\bibitem[von Marcard et~al.(2018)von Marcard, Henschel, Black, Rosenhahn, and Pons-Moll]{vonMarcard2018}
Timo von Marcard, Roberto Henschel, Michael Black, Bodo Rosenhahn, and Gerard Pons-Moll.
\newblock Recovering accurate 3d human pose in the wild using imus and a moving camera.
\newblock In \emph{European Conference on Computer Vision (ECCV)}, 2018.

\bibitem[Wang et~al.(2021{\natexlab{a}})Wang, Manhardt, Tombari, and Ji]{Wang_2021_CVPR}
Gu Wang, Fabian Manhardt, Federico Tombari, and Xiangyang Ji.
\newblock Gdr-net: Geometry-guided direct regression network for monocular 6d object pose estimation.
\newblock In \emph{Proceedings of the IEEE/CVF Conference on Computer Vision and Pattern Recognition (CVPR)}, pages 16611--16621, 2021{\natexlab{a}}.

\bibitem[Wang et~al.(2021{\natexlab{b}})Wang, Manhardt, Tombari, and Ji]{Wang_2021_GDRN}
Gu Wang, Fabian Manhardt, Federico Tombari, and Xiangyang Ji.
\newblock {GDR-Net}: Geometry-guided direct regression network for monocular 6d object pose estimation.
\newblock In \emph{IEEE/CVF Conference on Computer Vision and Pattern Recognition (CVPR)}, pages 16611--16621, 2021{\natexlab{b}}.

\bibitem[Wang et~al.(2019{\natexlab{a}})Wang, Sridhar, Huang, Valentin, Song, and Guibas]{Wang_2019_CVPR_NOCS}
He Wang, Srinath Sridhar, Jingwei Huang, Julien Valentin, Shuran Song, and Leonidas~J. Guibas.
\newblock Normalized object coordinate space for category-level 6d object pose and size estimation.
\newblock In \emph{The IEEE Conference on Computer Vision and Pattern Recognition (CVPR)}, 2019{\natexlab{a}}.

\bibitem[Wang et~al.(2019{\natexlab{b}})Wang, Sun, Cheng, Jiang, Deng, Zhao, Liu, Mu, Tan, Wang, Liu, and Xiao]{Wang19HRNet}
Jingdong Wang, Ke Sun, Tianheng Cheng, Borui Jiang, Chaorui Deng, Yang Zhao, Dong Liu, Yadong Mu, Mingkui Tan, Xinggang Wang, Wenyu Liu, and Bin Xiao.
\newblock Deep high-resolution representation learning for visual recognition.
\newblock \emph{TPAMI}, 2019{\natexlab{b}}.

\bibitem[Wen and Bekris(2021)]{wen_bundletrack_2021}
Bowen Wen and Kostas Bekris.
\newblock {BundleTrack}: {6D} {Pose} {Tracking} for {Novel} {Objects} without {Instance} or {Category}-{Level} {3D} {Models}.
\newblock In \emph{2021 {IEEE}/{RSJ} {International} {Conference} on {Intelligent} {Robots} and {Systems} ({IROS})}, pages 8067--8074, Prague, Czech Republic, 2021. IEEE Press.

\bibitem[Wen et~al.(2023)Wen, Tremblay, Blukis, Tyree, M\"{u}ller, Evans, Fox, Kautz, and Birchfield]{bundlesdfwen2023}
Bowen Wen, Jonathan Tremblay, Valts Blukis, Stephen Tyree, Thomas M\"{u}ller, Alex Evans, Dieter Fox, Jan Kautz, and Stan Birchfield.
\newblock {BundleSDF}: {N}eural 6-{DoF} tracking and {3D} reconstruction of unknown objects.
\newblock In \emph{CVPR}, 2023.

\bibitem[Xiang et~al.(2018)Xiang, Schmidt, Narayanan, and Fox]{xiang2018posecnn}
Yu Xiang, Tanner Schmidt, Venkatraman Narayanan, and Dieter Fox.
\newblock Posecnn: A convolutional neural network for 6d object pose estimation in cluttered scenes.
\newblock 2018.

\bibitem[Xie et~al.(2022)Xie, Bhatnagar, and Pons-Moll]{xie22chore}
Xianghui Xie, Bharat~Lal Bhatnagar, and Gerard Pons-Moll.
\newblock Chore: Contact, human and object reconstruction from a single rgb image.
\newblock In \emph{European Conference on Computer Vision ({ECCV})}. {Springer}, 2022.

\bibitem[Xie et~al.(2023{\natexlab{a}})Xie, Bhatnagar, Lenssen, and Pons-Moll]{xie2023template_free}
Xianghui Xie, Bharat~Lal Bhatnagar, Jan~Eric Lenssen, and Gerard Pons-Moll.
\newblock Template free reconstruction of human-object interaction with procedural interaction generation.
\newblock In \emph{ArXiv}, 2023{\natexlab{a}}.

\bibitem[Xie et~al.(2023{\natexlab{b}})Xie, Bhatnagar, and Pons-Moll]{xie2023vistracker}
Xianghui Xie, Bharat~Lal Bhatnagar, and Gerard Pons-Moll.
\newblock Visibility aware human-object interaction tracking from single rgb camera.
\newblock In \emph{IEEE Conference on Computer Vision and Pattern Recognition (CVPR)}, 2023{\natexlab{b}}.

\bibitem[Xiu et~al.(2022)Xiu, Yang, Tzionas, and Black]{xiu2022icon}
Yuliang Xiu, Jinlong Yang, Dimitrios Tzionas, and Michael~J. Black.
\newblock {ICON}: {I}mplicit {C}lothed humans {O}btained from {N}ormals.
\newblock In \emph{Proceedings of the IEEE/CVF Conference on Computer Vision and Pattern Recognition (CVPR)}, pages 13296--13306, 2022.

\bibitem[Xiu et~al.(2023)Xiu, Yang, Cao, Tzionas, and Black]{xiu2023econ}
Yuliang Xiu, Jinlong Yang, Xu Cao, Dimitrios Tzionas, and Michael~J. Black.
\newblock {ECON: Explicit Clothed humans Optimized via Normal integration}.
\newblock In \emph{Proceedings of the IEEE/CVF Conference on Computer Vision and Pattern Recognition (CVPR)}, 2023.

\bibitem[Xu et~al.(2019)Xu, Wang, Ceylan, Mech, and Neumann]{DISN}
Qiangeng Xu, Weiyue Wang, Duygu Ceylan, Radomir Mech, and Ulrich Neumann.
\newblock Disn: Deep implicit surface network for high-quality single-view 3d reconstruction.
\newblock In \emph{Advances in Neural Information Processing Systems}. Curran Associates, Inc., 2019.

\bibitem[Yang et~al.(2021{\natexlab{a}})Yang, Sun, Jampani, Vlasic, Cole, Chang, Ramanan, Freeman, and Liu]{yang2021lasr}
Gengshan Yang, Deqing Sun, Varun Jampani, Daniel Vlasic, Forrester Cole, Huiwen Chang, Deva Ramanan, William~T Freeman, and Ce Liu.
\newblock Lasr: Learning articulated shape reconstruction from a monocular video.
\newblock In \emph{CVPR}, 2021{\natexlab{a}}.

\bibitem[Yang et~al.(2022)Yang, Vo, Neverova, Ramanan, Vedaldi, and Joo]{yang2022banmo}
Gengshan Yang, Minh Vo, Natalia Neverova, Deva Ramanan, Andrea Vedaldi, and Hanbyul Joo.
\newblock Banmo: Building animatable 3d neural models from many casual videos.
\newblock In \emph{CVPR}, 2022.

\bibitem[Yang et~al.(2023)Yang, Wang, Reddy, and Ramanan]{yang2023rac}
Gengshan Yang, Chaoyang Wang, N.~Dinesh Reddy, and Deva Ramanan.
\newblock Reconstructing animatable categories from videos.
\newblock In \emph{CVPR}, 2023.

\bibitem[Yang et~al.(2021{\natexlab{b}})Yang, Zhan, Li, Xu, Li, and Lu]{yang2021cpf}
Lixin Yang, Xinyu Zhan, Kailin Li, Wenqiang Xu, Jiefeng Li, and Cewu Lu.
\newblock {CPF}: Learning a contact potential field to model the hand-object interaction.
\newblock In \emph{ICCV}, 2021{\natexlab{b}}.

\bibitem[Ye et~al.(2023)Ye, Hebbar, Gupta, and Tulsiani]{Ye_2023_ICCV}
Yufei Ye, Poorvi Hebbar, Abhinav Gupta, and Shubham Tulsiani.
\newblock Diffusion-guided reconstruction of everyday hand-object interaction clips.
\newblock In \emph{Proceedings of the IEEE/CVF International Conference on Computer Vision (ICCV)}, pages 19717--19728, 2023.

\bibitem[Yi et~al.(2022)Yi, Huang, Tzionas, Kocabas, Hassan, Tang, Thies, and Black]{Yi_MOVER_2022}
Hongwei Yi, Chun-Hao~P. Huang, Dimitrios Tzionas, Muhammed Kocabas, Mohamed Hassan, Siyu Tang, Justus Thies, and Michael~J. Black.
\newblock Human-aware object placement for visual environment reconstruction.
\newblock In \emph{IEEE/CVF Conf.~on Computer Vision and Pattern Recognition (CVPR)}, pages 3959--3970, 2022.

\bibitem[Yuan et~al.(2022)Yuan, Iqbal, Molchanov, Kitani, and Kautz]{yuan2022glamr}
Ye Yuan, Umar Iqbal, Pavlo Molchanov, Kris Kitani, and Jan Kautz.
\newblock Glamr: Global occlusion-aware human mesh recovery with dynamic cameras.
\newblock In \emph{Proceedings of the IEEE/CVF Conference on Computer Vision and Pattern Recognition (CVPR)}, 2022.

\bibitem[Zhang et~al.(2023)Zhang, Luo, Yang, Xu, Wu, Shi, Yu, Xu, and Wang]{zhang2023neuraldome}
Juze Zhang, Haimin Luo, Hongdi Yang, Xinru Xu, Qianyang Wu, Ye Shi, Jingyi Yu, Lan Xu, and Jingya Wang.
\newblock Neuraldome: A neural modeling pipeline on multi-view human-object interactions.
\newblock In \emph{CVPR}, 2023.

\bibitem[Zhang et~al.(2020)Zhang, Pepose, Joo, Ramanan, Malik, and Kanazawa]{zhang2020phosa}
Jason~Y. Zhang, Sam Pepose, Hanbyul Joo, Deva Ramanan, Jitendra Malik, and Angjoo Kanazawa.
\newblock Perceiving 3d human-object spatial arrangements from a single image in the wild.
\newblock In \emph{European Conference on Computer Vision (ECCV)}, 2020.

\bibitem[Zhang et~al.(2018)Zhang, Zhang, Zhang, Tenenbaum, Freeman, and Wu]{genre}
Xiuming Zhang, Zhoutong Zhang, Chengkai Zhang, Joshua~B Tenenbaum, William~T Freeman, and Jiajun Wu.
\newblock {Learning to Reconstruct Shapes From Unseen Classes}.
\newblock In \emph{Advances in Neural Information Processing Systems (NeurIPS)}, 2018.

\bibitem[Zhang et~al.(2022)Zhang, Bhatnagar, Starke, Guzov, and Pons-Moll]{zhang2022couch}
Xiaohan Zhang, Bharat~Lal Bhatnagar, Sebastian Starke, Vladimir Guzov, and Gerard Pons-Moll.
\newblock Couch: Towards controllable human-chair interactions.
\newblock In \emph{European Conference on Computer Vision ({ECCV})}. {Springer}, 2022.

\bibitem[Zhou et~al.(2022)Zhou, Bhatnagar, Lenssen, and Pons-Moll]{zhou2022toch}
Keyang Zhou, Bharat~Lal Bhatnagar, Jan~Eric Lenssen, and Gerard Pons-Moll.
\newblock Toch: Spatio-temporal object correspondence to hand for motion refinement.
\newblock In \emph{European Conference on Computer Vision ({ECCV})}. {Springer}, 2022.

\bibitem[Zhou et~al.(2019)Zhou, Barnes, Jingwan, Jimei, and Hao]{Zhou_2019_CVPR}
Yi Zhou, Connelly Barnes, Lu Jingwan, Yang Jimei, and Li Hao.
\newblock On the continuity of rotation representations in neural networks.
\newblock In \emph{The IEEE Conference on Computer Vision and Pattern Recognition (CVPR)}, 2019.

\end{thebibliography}
}

\end{document}


\maketitle

Supplementary material:
what is promised in the main paper:
\begin{enumerate}
    \item Implementation details: 
        \begin{itemize}
            \item blender rendering. 
            \item Our HDM network details. 
            \item Our \dataName{} data distribution. 
        \end{itemize}
    \item Experiments: 
    \begin{itemize}
        \item Analysis of $T_0$ for the second stage: done. Performance vs. diffusion step plot is done, consistent with our setup in the main paper. 
        \item More examples for generalization experiment. 
        \item Failure case, limitations and future works. 
    \end{itemize}
    \item What else can we add? (not mentioned in main, but can be added in case reviewers ask). 
    \begin{itemize}
        \item The reconstruction performance of our autoencoder. 
        \item Evaluate quality of our dataset: Contact distribution of our \dataName{} and original BEHAVE data, category wise. Draw contact heat maps on SMPL surfaces. 
    \end{itemize}
\end{enumerate}

\textbf{Summary of results to show:}
\begin{enumerate}
    \item Demonstrate our dataset: flying camera of our data arrays
    \begin{itemize}
        \item Start with a simple example of changing object given one interaction pose 
        \item After explaining the method, show flying camera video, infinite amount of interactions (maybe repeated but also try to get diverse) 
    \end{itemize}
    \item Demonstrate our method: 
    \begin{itemize}
        \item Show more reconstruction results from COCO and other datasets
        \item Show baseline results, PC2, PC2 + \dataName{}, and ours 
        \item Demonstrate that our method can do segmentation and it is important: 
            \begin{itemize}
                \item Show that we can manipulate the objects, e.g. scale the object and optimize while maintaining the contact
                \item Show that we can texturize human and object differently. Related works: Fantasia, ICCV'23 (Generate texture from mesh inputs). Text2txt: mesh to texture based on text input.  
            \end{itemize}
        \item Concerns that our method does not really reconstruct the full surface, i.e. as meshes. Try to use IF-Net to get meshes. Other related works: Delaunay triangulation, Shape as Points. 
    \end{itemize}
\end{enumerate}

{
    \small
    \bibliographystyle{ieeenat_fullname}
    \bibliography{main}
}